\documentclass[final,5p,times,twocolumn,authoryear]{elsarticle}

\usepackage{amssymb}
\usepackage{amsmath}
\usepackage{siunitx}
\usepackage{url}
\journal{arXiv}

\begin{document}
\begin{frontmatter}

\title{Reconstructing Nodal Pressures in Water Distribution Systems with Graph Neural Networks}

\author[HDS]{Gergely Hajgató\corref{hajg}}
\ead{ghajgato@hds.bme.hu}
\cortext[hajg]{Corresponding author.}
\author[TMIT]{Bálint Gyires-Tóth}
\author[HDS]{György Paál}

\affiliation[HDS]{organization={Department of Hydrodynamic Systems},
            addressline={Műegyetem rkp. 3.}, 
            city={Budapest},
            postcode={HU-1111}, 
            country={Hungary}}
\affiliation[TMIT]{organization={Department of Telecommunications and Media Informatics},
            addressline={Magyar tudósok körútja 2.}, 
            city={Budapest},
            postcode={HU-1117}, 
            country={Hungary}}

\begin{abstract}
A novel data-driven method to reconstruct nodal pressures in water distribution systems (WDSs) is introduced in this study.
For the safe and efficient operation it would be critical to know the nodal pressure distribution in a WDS, yet, it is infeasible due to the limited number of instrumentation.

This paper proposes a graph neural network-based (GNN) novel approach for nodal pressure recovery. 
An edge-weighting method is introduced in order to take hydraulic losses account in the graph convolutions.
The findings are evaluated on data generated from three benchmark WDSs with a realistic nodal demand variation.
The model performance is compared to interpolated regularization, a parameter-free graph signal reconstruction method.
The theoretical considerations of hyperparameter selection of the GNN are also discussed.

The proposed data-driven model is able to reconstruct the nodal pressure with at most $5\%$ relative error on average at an observation ratio of $5\%$ at least.
A further advantage of the GNN is that connections do not have to be weighted, as binary connections results in similar accuracy. This findings suggests, that the extra information \textit{hidden} in the data can be learned by the neural network.
Moreover, the results are superior to the analytical, interpolated regularization approach in each tested case.
\end{abstract}

\end{frontmatter}

\section{Introduction}
\label{sec:intro}
As data acquisition is indispensable for digital twins to ensure safe and efficient operation gathering real-time information of hydrodynamic properties is an essential task for water distribution system (WDS) operators.
Beside its importance, collecting and processing data from a WDS is difficult in two respects: the data lie on an irregular grid and they are only partially observable.

Digital twins\footnote{Digital twins are numerical models of physical processes occurring in an environment. Digital twins are engineered to be evaluated in real-time in order to analyze the ongoing processes in the specific environment.} and problem-tailored numerical models for WDS management is an actively researched area with a vivid research community.
Measurement-dependent models are successful in optimal pump scheduling \citep{Candelieri2020,Bagloee2018,Odan2014,Bhattacharya2003}, in water demand forecast \citep{Farias2018,Herrera2010} and in attack and leakage detection \citep{Taormina2018,Taormina2018a}, among others.
The efficacy of these models depends on the quality and quantity of live measurement data, mostly nodal pressures in the first place.

These demands catalyzed developments on online sensing, and these efforts, in conjunction with the available numerical models led to the birth of the smart water grid (smart water network, smart urban water network) paradigm \citep{Lee2014}.
Still, the high number of pipes and junctions in a real-life WDS \citep{Sitzenfrei2013,Yazdani2011} makes full observation of a hydraulic variable over the whole pipe network infeasible \citep{Klise2013}.
This induces the need for methods that can complete partially observed graph signals.

Graph signal reconstruction is a cornerstone of efficient operation of sensor networks as the operating costs decrease by using only a part of the installed sensors to recover the signal in every node.
Hence, the completion of partially observed signals and the optimal selection of sensors for effective signal reconstruction are often discussed together in the literature.

\citet{Tanaka2020} summarize the ongoing trends in graph signal sampling and reconstruction and discuss a popular approach of linear regression on graph signals.
Despite the advances in kernel-based regression methods \citep{Romero2017,Venkitaraman2019}, linear regression is favored in recent studies \citep{FerrerCid2021,Jiang2020} due to its non-parametric property.

The literature discussing the application of signal completion models in water distribution systems is limited.
\citet{Xie2017} developed an algorithm to optimally distribute the sensors over WDSs by minimizing the mutual coherence.
The main goal of their work was to detect leaks in the inspected WDSs.
\citet{Wei2020} introduce a novel graph Fourier-transform methodology to define optimal sensor placement for observing transient phenomena, namely pollutant spread.
The temporal signal is reconstructed perfectly by observing $30\%$ of the nodes, if the sensor deployment is done according to their algorithm.

A data-driven approach for nodal pressure reconstruction in real-time is presented by \citet{Lima2017}, who developed a model based on a multilayer perceptron (MLP) to complete partial nodal pressure signals.
The dataset is based on numerical simulations, where the nodal demands are the boundary conditions of the hydraulic model and their variation over a day is handled by changing all of the demands from scene-to-scene with a common multiplier.
This is a significant simplification of the problem as nodal demands do not necessarily change at once and with the same ratio in a WDS.
The MLP with one hidden layer with 10 neurons achieves an average reconstruction error below $1\%$ on two WDSs with approx. $150$ nodes each.

Although the example of a shallow MLP succeeded in the above simplified nodal pressure completion problem, using MLPs for larger water networks with more realistic nodal demand variation could be inefficient.
Recent advances of graph signal processing \citep{Shuman2013} led to various formulations of the convolution operator on graphs, making the learning efficient on irregular domains.
An in-depth survey on graph neural networks (GNNs) is given by \citet{Wu2021} and a benchmark of the popular methods is researched by \citet{Dwivedi2020}.

GNNs are applied in recent studies to solve WDS-related problems.
\citet{Garzon2021} find that their proposed GNN architecture is superior over MLPs in predicting the resilience of a WDS.
\citet{Tsiami2021} introduce a neural network architecture to localize cyber-physical attacks in WDSs.
The spatio-temporal signal is processed with neighborhood nodal feature aggregation in the spatial dimension and with 1D convolution on the temporal scale.
The original WDS topologies are transformed to a smaller, so-called \textit{condensed} network as the method does not handle situations where signal is missing from some of the nodes.
\citet{Zhou2021} proposes a convolution-based method for contamination source identification.
WDS topologies are segmented to smaller parts with fixed-sized adjacency matrices prior convolution in order to learn general filters.
The knowledge transfer is successful in the benchmark case where the learned filters are used on a bigger segment of the original WDS topology that was not seen during training.
The similarity between the smaller segments in the convolution process holds because pipe properties are not considered explicitly in the data.
The authors mention this as a possible drawback because the learned filters are general only in a limited domain where the ignored properties (e.g. pipe length and diameter) are similar among the segmented graphs.

Beside the above developments, a robust method for nodal pressure recovery and its validation on realistic scenarios has not been published yet in the literature.
This study covers this hiatus; thus, the methodology of applying graph neural networks to the nodal pressure reconstruction problem is discussed here.
Our contributions are the following:
\begin{itemize}
    \item a specific type of GNN is proposed according to the peculiarities of the graph representation of water networks;
    \item a way of weighting graph edges is presented based on hydraulic properties;
    \item a guide is given on selecting the appropriate GNN hyperparameters for nodal pressure recovery;
    \item a data generation method is presented with a realistic nodal-demand variation model;
    \item the signal reconstruction capability of the proposed GNN architecture with and without the proposed edge-weighting is studied on three benchmark WDSs;
    \item the signal reconstruction performance of the proposed model is compared to a na\"ive model and to a parameter-free linear regression model.
\end{itemize}
The paper is organized as follows. The proposed methodology of nodal pressure reconstruction with GNNs is discussed in the next section \textit{Nodal pressure reconstruction with graph neural networks}. The data generation method together with information on the model training are discussed in the subsequent section \textit{Experiments}. The results are evaluated and concluded thereafter in sections \textit{Evaluation and results} and \textit{Conclusions}, respectively.

\begin{figure*}
    \centering
    \includegraphics[width=\textwidth, keepaspectratio]{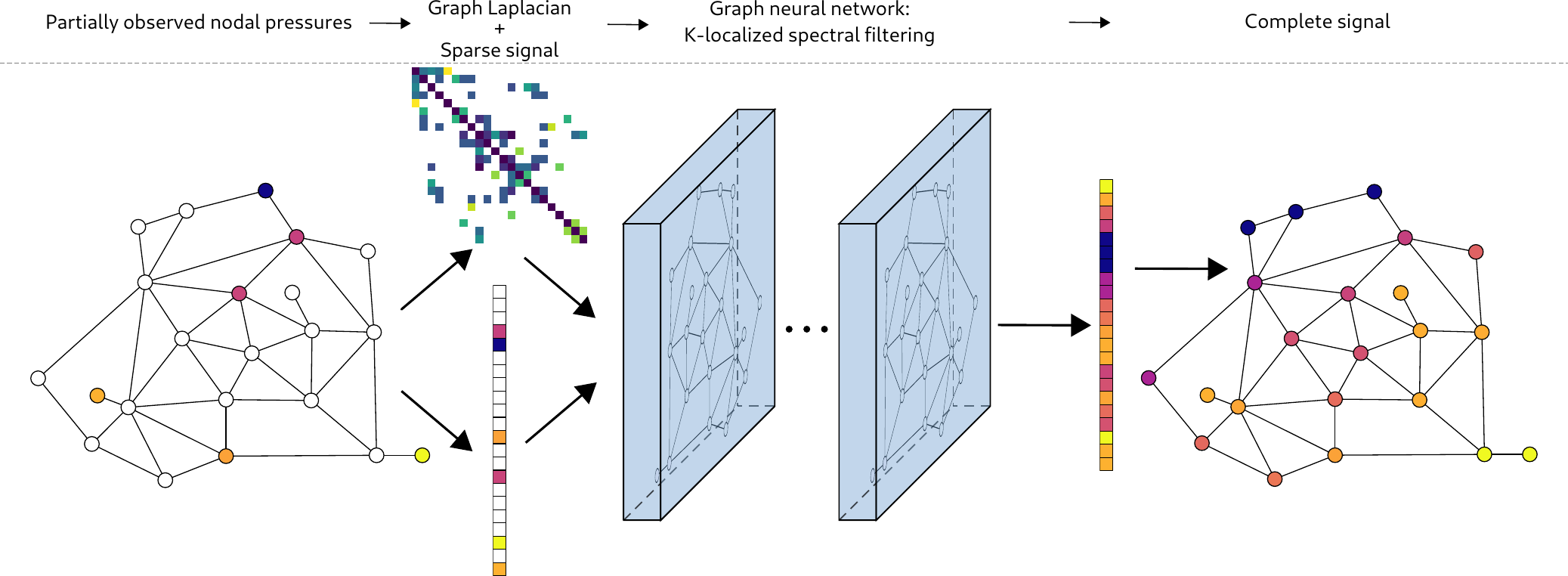}
    \caption{Overview of the proposed method for nodal pressure reconstruction. Pressure transducers are installed to the colored nodes, while empty nodes are unobserved. The color refers to the measured pressure (scale is not presented). The vector of the sparse pressure signal with the graph Laplacian is fed to the graph neural network that reconstructs the signal in the unobserved nodes.}
    \label{fig:overview}
\end{figure*}

\section{Nodal pressure reconstruction with graph neural networks}
\label{sec:method}
\subsection{Problem definition}
\label{sec:problem}
The pipe network of a water distribution system is modeled as a connected, undirected graph $G=(V,E)$ wherein the $V$ set of vertices represent pipe joints, consuming points, tanks; and the $E$ set of edges represent pipes, pumps and valves.
Such hydraulic topologies are depicted in Fig.~\ref{fig:topos}.
An adjacency matrix $A$ with weights $a_{ij}$ is associated the edges as follows.
Element $a_{ij}$ is $0$ if there is no connection between the $i\mathrm{th}$ and $j\mathrm{th}$ node; it is $1$ if there is a connection and the weighting is omitted from the adjacency matrix; and it is some real value if there is a connection and some weighting function is considered.
The possible methods for weighting the connections is discussed in Sec.~\ref{sec:effects}.

The $Y$ nodal pressure is a spatio-temporal signal distributed along the nodes.
Continuous and complete observation is infeasible due to the lack of sensors, i.e., pressure sensors are deployed only to the fraction of the nodes.
The allocation of the sensors is fixed; there are both observed and unobserved nodes in the WDS.
The task in this context is to observe the partial graph signal $Y^*$ and reconstruct the unobserved part $\hat{Y}$.

Transient phenomena (pressure transients) are not involved in the study as they are rare under normal circumstances and they are omitted in efficient operation planning either.
Thus, the nodal pressure distribution corresponding to the steady states are examined, which means that the signal has Markov property in time.
Moreover, the signal has pairwise Markov property spatially, but this feature is not exhausted by the studied methods.

\subsection{Modeling the hydraulics}
\label{sec:govern}
Nodal pressure reconstruction is related to fluid flow in pipe networks.
In order to find a suitable method for data-driven nodal pressure reconstruction, taking a look at the underlying physics and mathematical representation beforehand can reveal some valuable observations.

The governing equations of fluid flow in a pipe network are the continuity equation and the energy equation.
The law of continuity applied to the $l$-th node is
\begin{equation}
    \label{eq:conti}
    \sum_{j=1}^{f_l} \rho \cdot Q_{j,l} = \dot{m}_l\mathrm{,}
\end{equation}
where $f_l$ is the number of edges ending in the $l$-th node, $\rho$ is the density of the fluid, $Q_{j,l}$ is the signed volumetric flow rate flowing in and out of the node, and $\dot{m}_l$ is the nodal consumption as mass flow rate.
The energy equation with a loss-term applied on the $k$-th edge is
\begin{equation}
    \label{eq:energy}
    p_{k,I}-p_{k,II} + A_k \cdot Q_k + B_k \cdot Q_k^2 + C_k \cdot \lvert{Q_k}\rvert \cdot Q_k + D_k = 0\mathrm{,}
\end{equation}
where $p_{I}$ and $p_{II}$ are the nodal pressures in the ends of the edge, $Q_k$ is the volumetric flow rate and $A$, $B$, $C$ and $D$ are edge-dependent factors and constants.

Nodal pressure is considered to be a valuable information, hence Eq.~\ref{eq:energy} describes the way, how information propagates in the water network.
Considering a pipe from a real world WDS, the following assumptions can be made.
\begin{itemize}
    \item $A_k \cdot Q_k = 0$. Assuming a high Reynolds number flow, this term is negligible compared to quadratic loss term.
    \item $B_k \cdot Q_k^2 = 0$ assuming constant-diameter pipes, as the quadratic term is the difference between the dynamic pressures at the ends of the $k$th pipe.
\end{itemize}
$D_k$ comes from the geodetic height difference, hence, it is a constant term, and can be embedded in the pressure terms directly.
$C_k \cdot \lvert{Q_k}\rvert \cdot Q_k$ describes the wall friction and separation losses in the $k$th pipe or other hydraulic component that is proportional to the square of the volumetric flow rate.
This term has a sign as it reduces the total pressure in the direction of the fluid flow.

This way, the energy equation with a loss-term can be simplified to
\begin{equation}
    \label{eq:energy-pipe}
    p_{k,I}^*-p_{k,II}^* + C_k \cdot \lvert{Q_k}\rvert \cdot Q_k = 0
\end{equation}
for a pipe of constant diameter which is the most common edge-like part in a water distribution system.
The term $D_k$ is embedded in the pressure terms denoted with a star, meaning, that $p_{k,I}^*$ and $p_{k,II}^*$ are containing the pressure of the fluid column according to the geodetic height of the respective node.
The direction of the edges can be arbitrary as the direction together with the sign of the volumetric flow rate tells whether the fluid flows from node $I$ to node $II$ or vice-versa.
Losses arising from other components and in the nodes are considered as part of the pipe-losses.

In order to get the hydraulic state of the water distribution system, both Eq.~\ref{eq:conti} and Eq.~\ref{eq:energy} have to be solved for all the node-like and edge-like elements of the WDS, respectively.
These equations form a system of nonlinear, algebraic equations that can be linearized by taking some further assumptions.
Knowing the boundary conditions (the nodal demands in practice), the system of the linearized equations can be solved by iterative numerical methods.
The hydraulic model was presented here in order to get a view on the underlying physics.
For the purposes of signal reconstruction, only the effect of the loss term will be examined in Sec.~\ref{sec:effects}.
The exact equations that were solved by the hydraulic simulator can be found in the EPANET 2 users manual \citep{Rossman2000}.

\subsection{Choosing the appropriate graph neural network}
\label{sec:chebnet}
Regarding the high number of existing graph neural network types \citep{Wu2021}, several aspects were considered taken before selecting the appropriate one.
Our main considerations were the following.
\begin{enumerate}
    \item The topology of the WDS is static. It is true for most of the time when the WDS operates under normal conditions. It does not hold, however, in emergency cases, when a segment of the WDS has to be shut off. Handling of such emergency scenarios is not considered in this paper.
    \item The input and output variables are nodal properties. This stands, as the task is to predict all the nodal pressures from a subset of the nodal pressures. The only other hydraulic property (a value proportional to hydraulic loss) is embedded in the adjacency matrix in some cases and neglected in others. No quantity is inferred on the edges nor added to the input data.
    \item The connection strength between nodes can be derived from the hydraulic properties, because a factor that is proportional to hydraulic loss coefficient can be computed before the training if the friction coefficient $C_k$ in Eq.~\ref{eq:energy-pipe} is available.
\end{enumerate}
Considering the above assumptions and properties, the K-localized spectral graph filtering method (referred later as ChebNet) was selected for the proposed model shown in Fig.~\ref{fig:overview}.
ChebNet was introduced in \citet{Defferrard2016}; only the core concept is described here.

As convolutional filters are efficient in learning data that are represented on a regular grid, \citet{Defferrard2016} formulated a similar filtering method for data represented on irregular grids.
ChebNet builds upon the fact that the convolutional operator is a single multiplication in the spectral domain.
Graph signals can be transformed to the spectral domain by the Laplace-transformation.
The transformation matrix can be calculated purely from topological data: the degree matrix $D$ and the adjacency matrix $A$.
The multiplication of this transformation matrix with the graph signal leads to the spectrum of the signal in one step.
This transformation is non-parametric and non-localized, which leaves no room for constructing convolutional filters.
\citet{Defferrard2016} propose the use of the truncated Chebyshev-polynomial of the graph Laplacian instead.
In this way, the range of information propagation can be controlled by the truncation.
Besides, different ranges can be parametrized differently, which allows to learn localized filters on the graph.

In practice, the scaled and normalized form of the Laplacian $\hat{L}$ is used instead of $L$ to ensure numerical stability:
\begin{equation}\label{eq:scaled_laplace}
    \hat{L} = \frac{2}{\lambda_{\mathrm{max}}}\cdot(I-{D^{-0.5}}\cdot{A}\cdot{D^{-0.5}}) - I\mathrm{,}
\end{equation}
where $\lambda_{\mathrm{max}}$ is the maximum of the eigenvalues of the graph Laplacian, $I$ the identity matrix, $D$ the degree matrix and $A$ the adjacency matrix.
The convolution of a weighted filter $g_{\Theta}$ with a graph signal $x$ becomes
\begin{equation}\label{eq:chebnet}
    g_{\Theta} * x = \sum_{k=1}^{K}{{\Theta}_{k} \cdot T_{k}(\hat{L}) \cdot x}\mathrm{,}
\end{equation}
where the Chebyshev-polynomials $T_k$ are of the first kind, thus these are in the form
\begin{equation}\label{eq:cheb_poly}
\begin{split}
    T_{0} &= I\mathrm{,}\\
    T_{1} &= \hat{L}\mathrm{,}\\
    T_{k \geq 2} &= 2 \cdot \hat{L} \cdot T_{k-1} - T_{k-2}\mathrm{.}
\end{split}
\end{equation}
The learnable parameters are denoted by $\Theta_k$.

Knowing the underlying physics and the learning model, the effects of the hyperparameters can be examined.

\subsection{Effects of ChebNet hyperparameters to nodal pressure reconstruction}
\label{sec:effects}
\subsubsection{Polynomial degree of the kernel}
The Chebyshev-polynomial of the graph Laplacian is taken into account up to the $K$-th order in  Eq.~\ref{eq:chebnet}.
The parameter $k$ sets how far the information propagates in one layer.
Several applications depend on information from the direct neighborhood only, and this property was exploited by \citet{Kipf2016} by truncating the Chebyshev-polynomial to the first degree and setting the localization by the number of consecutive layers.
Despite that, such an intensive truncation would not be beneficial for the current task if the number of the observed nodes are to be kept low.
From this perspective, the order of the polynomial is determined by the demand that the observed information reaches all the unobserved nodes.
This implies that -- as long as the observation points are selected uniformly randomly in the graph -- \textit{the polynomial order should be at least equal to the graph diameter}.

\subsubsection{Number of layers and channels}
Adding multiple channels to a layer allows the network to learn different filters for the same receptive field.
It will be necessary in this study as the underlying physics cannot be described from the partial observations with a single formula.

The number of layers plays a similar role in the information propagation as the order of the polynomial degree and the number of channels together.
The output of a ChebNet layer is constructed with a receptive field, characterized by the polynomial degree of the kernel as seen in Eq.~\ref{eq:chebnet}.
Stacking ChebNet layers consecutively, the information spreads not only due to powering the graph Laplacian but due to feeding the spread information to the input of the consecutive layers.

Sparse observation in large water networks would require a vast amount of hidden layers owing to the conclusion of the preceding section.
This is disadvantageous because of the vanishing gradient problem \citep{Hochreiter2001} that can lead to the failure of the model training.
Thus, the \textit{the expressive power required by the problem is ensured by the number of channels, while the number of layers is tried to be kept low} in the present study.

\subsubsection{Weighting of the graph Laplacian}
\begin{figure*}
    \centering
    \begin{tabular}{@{}c@{}}
        \includegraphics[width=.34\linewidth,keepaspectratio]{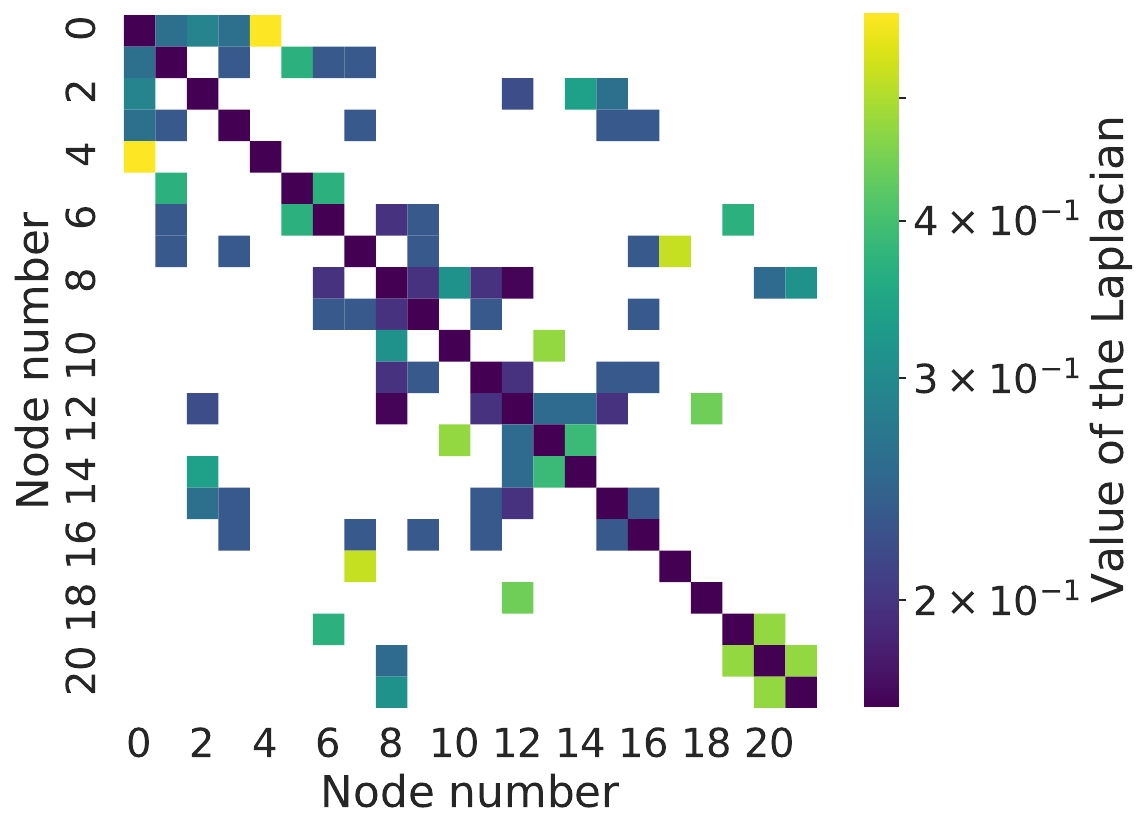} \\
        \footnotesize (a) Binary connections
    \end{tabular}
    \begin{tabular}{@{}c@{}}
        \includegraphics[width=.31\linewidth,keepaspectratio]{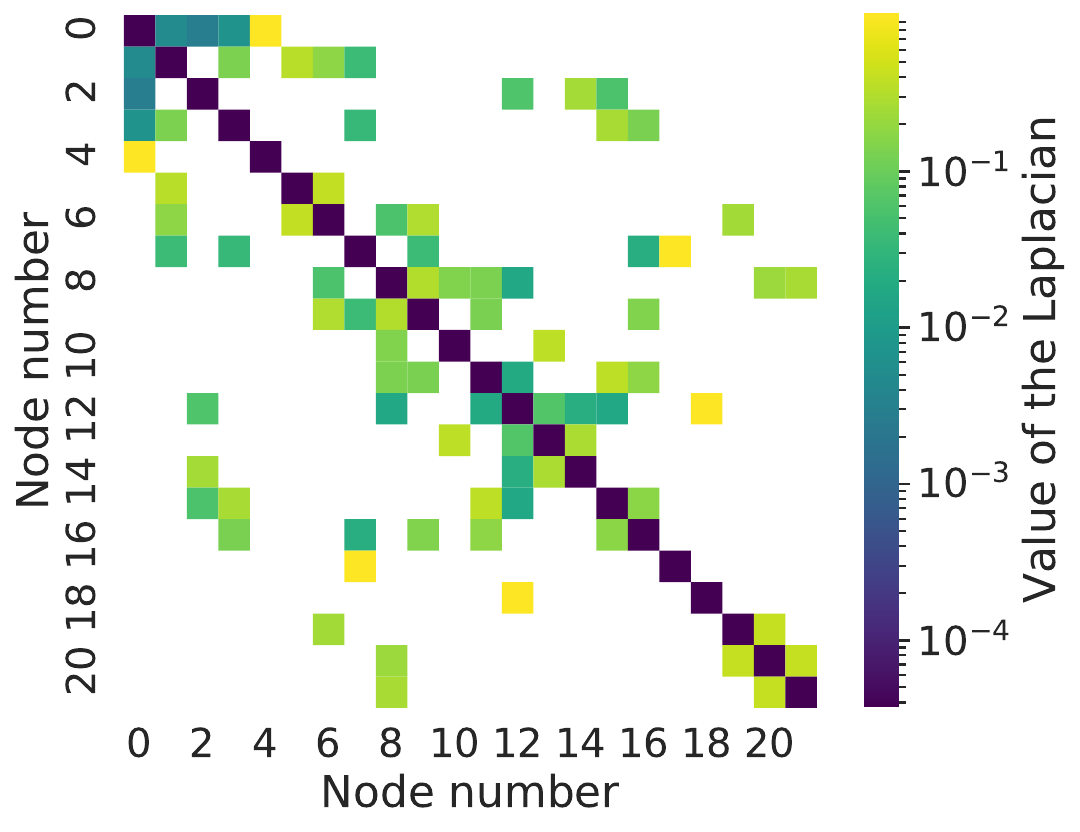} \\
        \footnotesize (b) Weighted connections
    \end{tabular}
    \begin{tabular}{@{}c@{}}
        \includegraphics[width=.34\linewidth,keepaspectratio]{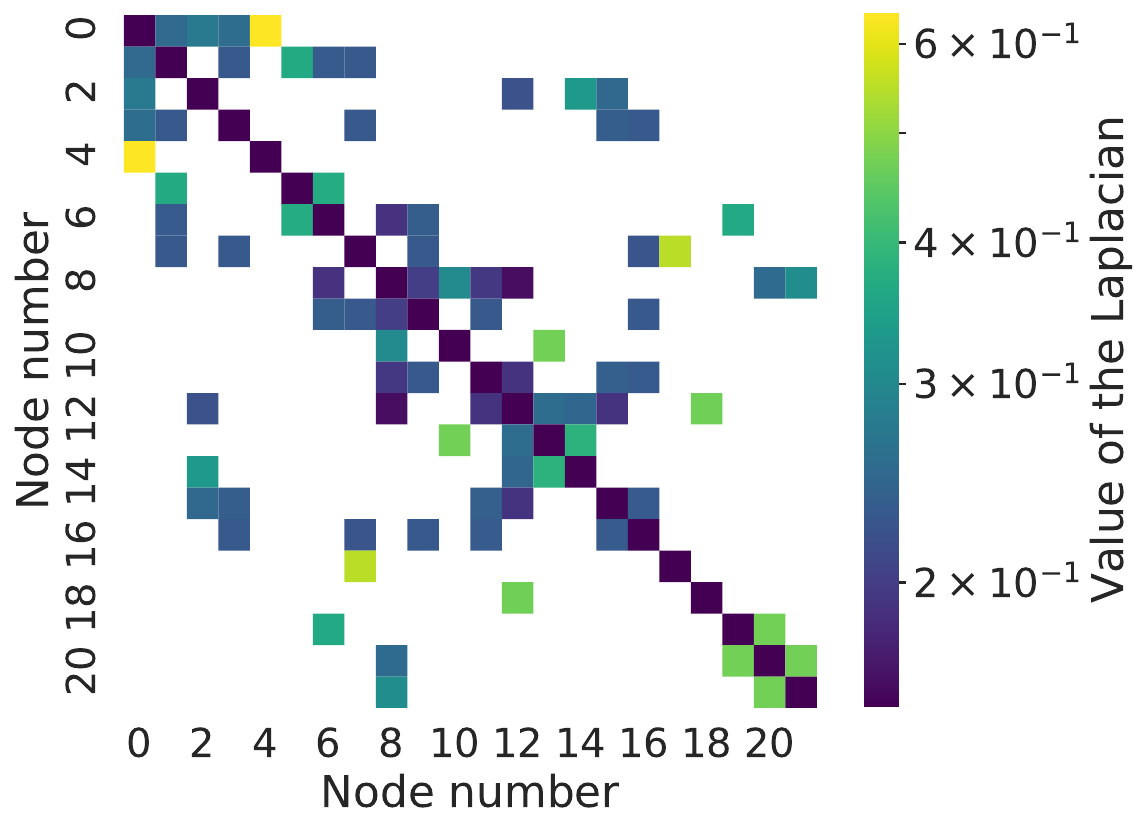} \\
        \footnotesize (c) Logarithmically weighted connections
    \end{tabular}
    \caption{Scaled and normalized graph Laplacian of Anytown calculated from the adjacency matrix with binary (a), weighted (b) and logarithmically weighted (c) connections.}
    \label{fig:graph_laplacians}
\end{figure*}

Most of the original work on graph convolution networks use the binary adjacency matrix to compute the graph Laplacian.
This is appropriate in many use cases where the connection strength between nodes is binary by its nature, e.g. in citation networks and knowledge graphs.

On the other hand, the connection between nodes in a water network can be considered weighted wherewith the learning could be preconditioned.
The question arises whether the prediction accuracy can be improved with this information embedded into the spectral filtering.
In other words, the weighting could indirectly lead to an anisotropic filtering (through the graph Laplacian) while using the isotropic filtering method (ChebNet).

Considering Eq.~\ref{eq:energy-pipe}, the first term on the right-hand side is
\begin{equation}
    \label{eq:hydraulic_loss_1}
    \Delta{p} = C_k \cdot Q_k^2
\end{equation}
by neglecting the direction of the flow.
As we are not interested in the flows on the edges, this is a reasonable simplification in order to get a quantity that is proportional to the resistance of information propagation between nodes.
For cylindrical pipes Eq.~\ref{eq:hydraulic_loss_1} becomes the Darcy--Weisbach equation:
\begin{equation}
    \label{eq:hydraulic_loss_2}
    \Delta{p} = \lambda \cdot l \cdot \frac{\rho \cdot Q^2}{8 \cdot d^5 \cdot \pi^2}\mathrm{,}
\end{equation}
where $\lambda$ is the friction coefficient, $l$ the length of the pipe, $\rho$ the fluid density and $d$ the diameter of the pipe.
Eq.~\ref{eq:hydraulic_loss_2} is untractable in many real-life situations as the friction coefficient is unknown.
Hence, it is a common practice to use empirical formulas instead.
Among these equations, the Hazen-Williams (H--W) formula is still favored despite its fairly inaccurate results.
The limitations of the formula are discussed thoroughly by \citet{Liou1998} and \citet{Christensen2000}.
Despite its drawbacks, the H--W formula is used in the present study due to the available data and the fact, that the applied hydraulic simulator used the formula for calculating the pressure loss.

The energy loss of the fluid flowing through a pipe according to the H--W formula is
\begin{equation}
    \label{eq:hazen-williams}
    H' = 4.727 \cdot C^{-1.852} \cdot d^{-4.871} \cdot l \cdot Q^{1.852}
\end{equation}
as per \citet{Rossman2000}.
Pipe diameter $d$ and pipe length $l$ are in feet and the volumetric flow rate $Q$ is in cubic feet per second.
$H'$ is in foot water column and called head loss.
$C$ is the Hazen--Williams roughness coefficient that is the attribute of the pipe and is defined empirically.
In practice, this $C$ coefficient is known for specific pipe materials while they are new.
The value in case of an aged pipe is only a crude guess as the roughness coefficient is influenced by a high number of external factors that cannot be measured.
Still, if the $C$ coefficient is known, the weight referring to how well the information (pressure) propagates in the water network can be derived from Eq.~\ref{eq:hazen-williams} as
\begin{equation}
    \label{eq:edge-weight}
    W = 4.727 \cdot \frac{C^{1.852} \cdot d^{4.871}}{l}\mathrm{.}
\end{equation}
This quantity is normalized by its maximum value calculated over the entire topology to avoid numerical problems if an extremely short or long pipe is presented in the WDS.
A more precise weighting metric can be defined from Eq.~\ref{eq:hydraulic_loss_2} in cases, where the flow coefficient $\lambda$ is tractable for all the pipes.

A key property of the scaled and normalized graph Laplacian calculated from the weighted adjacency matrix is that it spans a significantly wider range of values than the one calculated from the binary adjacency matrix.
The scaled and normalized graph Laplacians of the smallest water network involved in this study are depicted in Fig.~\ref{fig:graph_laplacians}.
These matrices were calculated from the binary, the weighted and the logarithmically weighted adjacency matrix, respectively.
Observing the figures, the self-connections lose their importance when the connections are weighted.
This is impractical for completing partial observations as self-connections are important in the nodes wherein a sensor is installed.
Moreover, self-connections help achieve better prediction accuracy in general, as shown by \citet{Wu2019}.

To overcome this problem and maintain the weighted nature of the adjacency matrix, the logarithm base 10 of the original weights are also used to weight the interconnections.
The connection represented by pumps and valves are weighted with connection strength $1$ in this case because all the valves in the numerical models were fully opened.
If partially closed valves are present in a WDS, a weighting can be defined specifically for valves from the energy equation similarly to the weights of the pipes.
In the case of pumps, a weight could be defined in a more elaborate way which is not discussed here as the number of pumps in a WDS is negligible compared to the number of pipes.

The scaled and normalized graph Laplacian according to the logarithmic weighting is depicted in Fig.~\ref{fig:graph_laplacians}.
The result is similar to that calculated from the binary adjacency matrix but with the connection strength encoded.

\textit{The effect of weighting and logarithmically weighting the connections with the plain binary connections will be examined in Sec.~\ref{sec:results}.}

\section{Experiments}
\label{sec:exps}
\subsection{Dataset}
\label{sec:setup}
The capability of graph neural networks to reconstruct partially observed graph signals were examined with numerical experiments.
The application is the reconstruction of all the nodal pressures in water networks from a limited number of measurement data.
The placement of the sensors is considered to be fixed, hence, a GNN is trained for a specific layout of the measurement devices.
Moreover, the nodal pressures are considered to be varying slowly in time (pressure waves may not developed) and temporality is omitted from the reconstruction model.
Instead, the model handles a scene at once, where a scene represents a steady state of the WDS according to the boundary conditions.
Pressure data collected from specific nodes per scene is given to the input of the model and all the nodal pressures are yielded as the output for the same scene.

Three, freely accessible benchmark WDSs were used for the numerical experiments to facilitate the reproduction of the results and the comparison with other models.
The dependence of the nodal pressures on the hydraulic state (nodal demands, pump speeds, etc.) of the WDS makes the hydraulic simulation necessary for every scene.
The numerical model was available for each of the water networks wherewith a sufficient amount of different scenes were generated to train, validate and test the proposed model.

\begin{figure}
    \centering
    \begin{tabular}{@{}c@{}}
        \includegraphics[width=\linewidth]{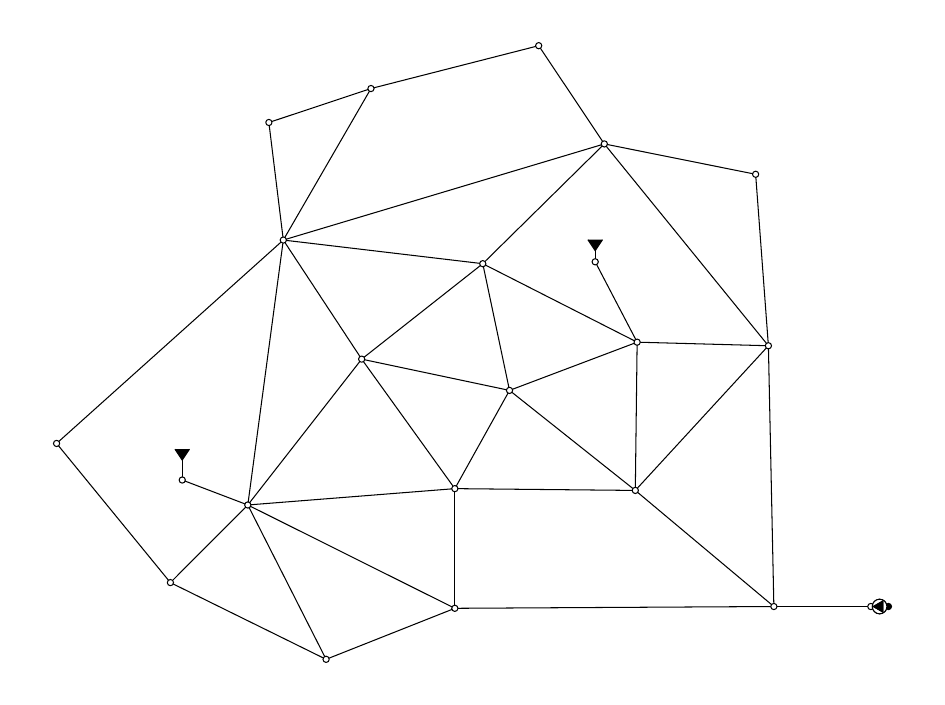}\\
        \footnotesize (a) Anytown
    \end{tabular}
    \begin{tabular}{@{}c@{}}
        \includegraphics[width=\linewidth]{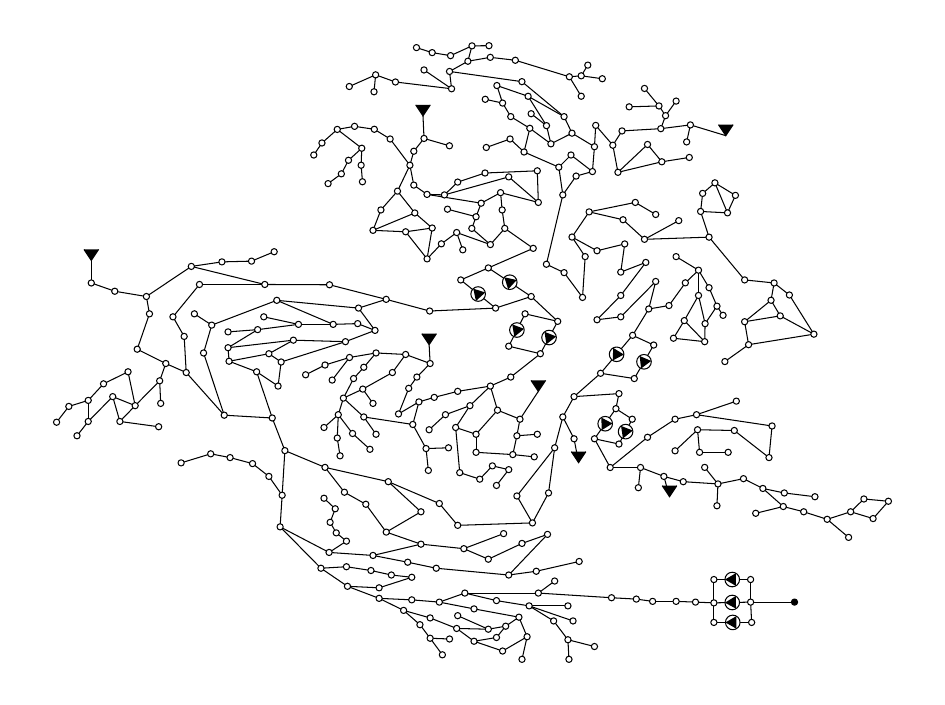}\\
        \footnotesize (b) C-Town
    \end{tabular}
    \begin{tabular}{@{}c@{}}
        \includegraphics[width=\linewidth]{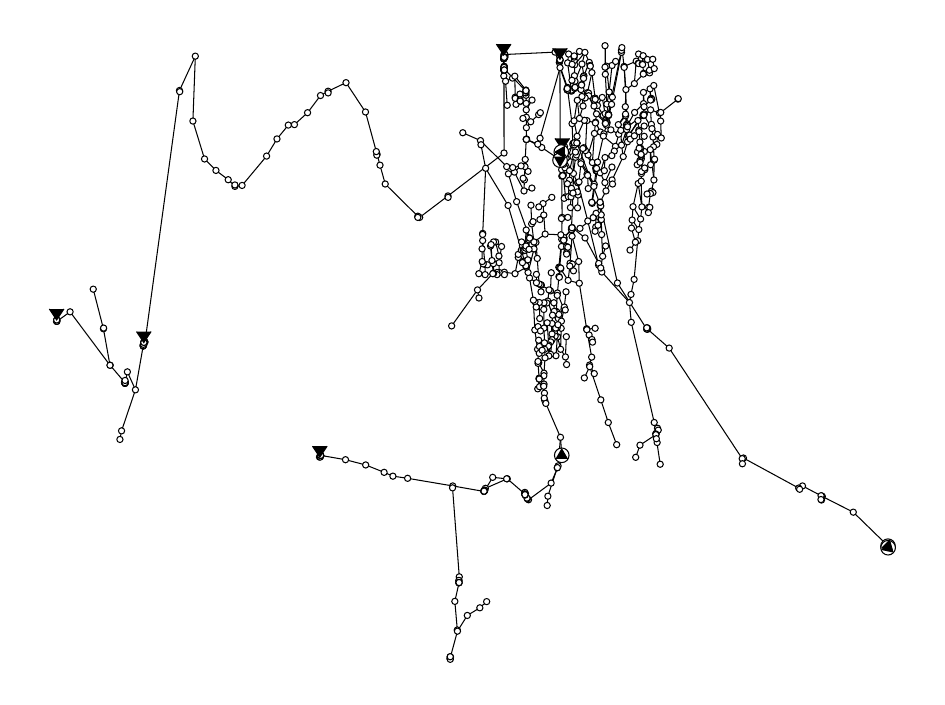}\\
        \footnotesize (c) Richmond
    \end{tabular}
    \caption{Topologies of the water distribution systems containing pipes, junctions, pumps and tanks.}
    \label{fig:topos}
\end{figure}

The water networks were namely the Anytown, C-Town and the Richmond networks.
These WDSs are commonly used benchmark networks among the water resources management community and can be reached online from the web page of the \citet{cwsExeter}.
Major properties of the water networks can be found in Table~\ref{tab:wds}, while the topologies of the networks are depicted in Fig.~\ref{fig:topos}.
Anytown and C-Town are fictitious villages, while Richmond is a town with around 8000 inhabitants in the United Kingdom.
The size of the water networks is reflected in the computation demand of the hydraulic simulations, therefore Anytown is involved in the study to examine the assumptions taken in Sec.~\ref{sec:effects}, while C-Town and Richmond represents the performance of the proposed model for real size WDSs.
Data on larger water networks were not available for the study.

\begin{table}[ht]
    \centering
    \begin{tabular}{l S[table-format=3.2] S[table-format=3.2] S[table-format=3.2]}
        \hline
        \textbf{Water network} & \textbf{\# of junctions} & \textbf{\# of pipes} & \textbf{Diameter}\\
        \hline
        Anytown & 22  & 41  & 5     \\
        C-Town  & 388 & 429 & 66    \\
        Richmond& 865 & 949 & 234   \\
        \hline
    \end{tabular}
    \caption{Major properties of the water distribution systems}
    \label{tab:wds}
\end{table}

Scene generation with varying consumer demands and pump speeds was conducted with the algorithm described by \citet{Hajgato2020}.
This method handles the day-night shift and the variation in the demands of the consumers in different phases thus a wide range of scenes can be generated that is simulating the variations in a real-life WDS.
A brief overview of the method is as follows.

First, the nodal consumption is generated for each node based on the initial data from the numerical model.
This initial value is an average for the consumption during the working hours.
It is assumed that the ratio of the water consumption between small (e.g. residential homes) and large (e.g. factories) consumers vary moderately in time, thus the initial demand is multiplied with random factors from a truncated normal distribution with the lower and upper limit aligned to the specific WDS.
The sum of the initial demands is rescaled thereafter to reflect the day-night shift in the water consumptions.
This is done by a multiplication with a factor from a uniform distribution and the lower and upper limits are WDS-specific again.
The nodal demands are then rescaled once more to satisfy the total consumption of the WDS.
Finally, pump speeds are generated between the feasibility limits with a uniform distribution.

$1000$, $10000$ and $20000$ different scenarios were generated with the above algorithm for the Anytown, C-Town and Richmond WDS, respectively.
The whole parameter space -- incl. total consumption, nodal consumptions and pump speeds -- were sampled with a design of experiments (DOE) method.
Although benefits are reported on the use of DOE methods in neural network training \citep{Mehta1999}, in the present study, latin hypercube sampling was used only to ensure that no repetitions occur in the database.

The generated scenes were split into training, validation and test data sets as per a ratio of $0.6$, $0.2$ and $0.2$, respectively.

\subsection{Baseline models}
\label{sec:baselines}
Two baseline models are presented to measure the performance of the proposed method and justify the data-driven approach of the task.

The first model is a na\"ive one with a minimal hypothesis set: all the missing nodal pressures are approximated with the mean of the known pressures.
\begin{equation}
    \label{eq:baseline-1}
    \hat{Y} = \frac{1}{k} \cdot \sum_{i=1}^k Y^*_i\mathrm{,}
\end{equation}
where $k$ denotes the number of observed nodes along the graph.

The second baseline model is the interpolated regularization which is the special case of the Tikhonov regularization on graph signals introduced by \citet{Belkin2004}.
The non-parametric nature of the model makes it favorable even in recent studies where it performs well among more recent signal recovery methods \citep{FerrerCid2021,Jiang2020}.
Most of the interpolation methods -- including interpolated regularization -- assume that the complete signal on the graph is smooth according to the metric in Eq.~\ref{eq:smoothness}.
\begin{equation}
    \label{eq:smoothness}
    \frac{1}{2}\sum_{i \approx j}A_{ij}\|Y_i-Y_j\|^2=\mathrm{tr}(Y^T L Y),
\end{equation}
where $L$ is the graph Laplacian.
\citet{Belkin2004} mentions several other domains based on the graph Laplacian where the smoothness can be measured, stating, that $L^2$ could be also a good choice for real-world problems.
Now, the signal completion task can be formulated considering that the complete signal \textit{has to be} smooth.
\begin{equation}
    \label{eq:intreg-opti}
    \min_{y}\:\mathrm{tr}(Y^T L Y)\quad\mathrm{s.t.}\:Y_i = Y_i^*, \:i={1..k},
\end{equation}
where node numbering is ordered such that the first $k$ elements belong to the observed nodes.
The first step of the interpolation is to rearrange the graph Laplacian (or some similar matrix representing the signal smoothness) in a way, that the first $k$ rows and columns represent the observed nodes.
This way, the Laplacian can be partitioned like in Eq.~\ref{eq:partition}.
\begin{equation}
    \label{eq:partition}
    L = \begin{pmatrix} L_1 & L_2 \\ L_2^T & L_3 \end{pmatrix},
\end{equation}
where $L_1$ is a $k \times k$ matrix belonging to the observed nodes, $L_2$ is $k \times (n-k)$ and $L_3$ is a $(n-k) \times (n-k)$ matrix belonging to the unobserved nodes only.
The signal in the unobserved nodes can be recovered like
\begin{equation}
    \label{eq:recon}
    \hat{Y} = S_3^{-1} S_2^T (Y^* + \mu \mathbf{1}),
\end{equation}
where $\mathbf{1}$ is a column vector of ones with a size $n-k$ and
\begin{equation}
    \label{eq:mu}
    \mu = - \frac{\sum_i^{n-k} {S_3^{-1} S_2^T Y^*}}{\sum_i^{n-k} {S_3^{-1} S_2^T \mathbf{1}}}.
\end{equation}
Although the orthogonality condition $Y \perp \mathbf{1}$ is not a strict one and it can be omitted, interpolated regularization was found to yield better results in the current experiments this way.

\subsection{Graph neural network training}
\label{sec:traineval}
Separate graph neural networks were built to signal reconstruction as per the considerations emphasized in Sec.~\ref{sec:effects}.
First, the GNN topology for the smallest WDS was defined by hyperparameter optimization discussed in Sec.~\ref{sec:ho}, then the resulting topologies were defined arbitrarily.
This compromise was taken due to limited resource availability.

The input and output were built from the previously generated dataset as follows.
Metrics were calculated from the training dataset for the standardization and normalization of the nodal pressures for feeding to the input and to the output of the GNNs, respectively.
Installing the sensors in the WDS was simulated by generating a binary mask with the same length as the number of nodes in the WDS.
Binary masks were generated randomly in the beginning of each training, according to a probability called \textit{observation ratio}.
The number of installed pressure sensors are defined by this ratio compared to the number of all nodes in the WDS.

The GNNs received an input with two channels: one channel for the partially observed signal (the full signal multiplied with the mask) and one channel for the mask itself.
The full signal was represented on the output of the GNN.

The training was carried out $5 \cdot 20 = 100$ times for every water network.
Five different observation ratios were examined, namely: ${0.05, 0.1, 0.2, 0.4, 0.8}$.
$20$ trainings were carried out with each observation ratio with different binary masks each to simulate the effect of different sensor placements.

The reconstruction capability of the model is measured by the mean-squared error between the model output and the ground truth during the training.
This loss is calculated for each node regardless whether it is an observed or an unobserved one, because preliminary studies showed better results with incorporating the observed nodes in the backpropagation of error.
The loss metric was measured on the training data and on the validation data at the end of every epoch, while it was measured only once on the test dataset at the end of every training.

\subsection{Hyperparameter optimization and network topologies}
\label{sec:ho}
The considerations presented in Sec.~\ref{sec:effects} served as an initial guess on the adequate neural network topology which was then modified via hyperparameter optimization in the case of the Anytown WDS.
A fairly shallow artificial neural network was initialized exclusively with ChebNet layers and weight decay as regularization.
Hidden layers utilized a sigmoid linear unit (SiLU) as activation \citep{Hendrycks2016}, while the output layer was activated by the sigmoid function to guarantee a finite-range feedback to the input signal.
Layer weights were initialized by the Xavier initialization \citep{Glorot2010}, while biases were initialized to zero.
As no experience has been gathered on the optimal learning rate of the problem, the Adam optimizer \citep{Kingma2015} was used for training both of the neural networks as it has an adaptive learning rate with momentum.

The number of the hidden layers, the number of the polynomial order of the kernel, the number of filters in a layer, the amount of regularization applied on each layer and the type of the adjacency matrix were subject to hyperparameter optimization as per Table~\ref{tab:ho}.

The tree-structured parzen estimator (TPE) \citep{Bergstra2011} algorithm was applied to find the optimal hyperparameters of the GNN.
TPE was allowed to do a search in the parameter space for $200$ evaluations and $50$ initial random trials were performed beforehand.
As multiple stochastic processes are included in the training, TPE was set to evaluate a specific hyperparameter setting as the average result of $5$ trainings.
The quality of a hyperparameter setting was measured by the validation error.
The training of a GNN was shut down after $2000$ epochs or earlier, if the validation error did not improve by more than \num{1e-6} in $50$ consecutive epochs.
In most of the cases, the training was ended before it reached the upper limit of epochs, thus raising the limit was not considered.

\begin{table}[htbp]
    \centering
    \begin{tabular}{l l l}
        \hline
        \textbf{Hyperparameter} & \textbf{Possibilities}\\
        \hline
        \# of layers & 2..4 \\
        $K_i$ & 30..50 \\
        $F_i$ & 30..50 \\
        weight decay & 1e-6..1e-4 \\
        Laplace weighting & \{binary, weighted, logarithmic\} \\
        \hline
    \end{tabular}
    \caption{Subjects of the hyperparameter optimization. $K_i$: degree of the Chebyshev-polynomial at the $i$th layer, $F_i$: number of filters in the $i$th layer.}
    \label{tab:ho}
\end{table}

The results of the hyperparameter optimization are shown in Section~\ref{sec:results}; only the final topologies of the GNNs are presented here.
The topology for the Anytown WDS was the direct outcome of the hyperparameter optimization.
The other two topologies were constructed arbitrarily but in concordance with the consequences of hyperparameter optimization.
The resulting GNN topologies are summarized in Table~\ref{tab:topo}.

\begin{table}[htbp]
    \centering
    \begin{tabular}{c S[table-format=5.0] S[table-format=6.0] S[table-format=6.0]}
        \hline
        \textbf{Hyperparams.} & \textbf{Anytown} & \textbf{C-Town} & \textbf{Richmond} \\
        \hline
        $K_1$ & 39  & 200   & 240   \\
        $K_2$ & 43  & 200   & 120   \\
        $K_3$ & 45  & 20    & 20    \\
        $F_1$ & 14  & 60    & 120   \\
        $F_2$ & 20  & 60    & 60    \\
        $F_3$ & 27  & 30    & 30    \\
        \hline
    \end{tabular}
    \caption{Graph neural network topologies per layer. $K_i$: degree of the Chebyshev-polynomial at the $i$th layer, $F_i$: number of filters in the $i$th layer.}
    \label{tab:topo}
\end{table}

\section{Evaluation and results}
\label{sec:results}
\subsection{Validation of the domain-specific considerations}
The domain-specific consideration taken in Sec.~\ref{sec:effects} are validated with hyperparameter optimization.
Results yielded by the various adjacency matrices and neural network depths are discussed in this section.

\begin{figure}[ht]
    \centering
    \begin{tabular}{@{}c@{}}
        \includegraphics[width=\linewidth]{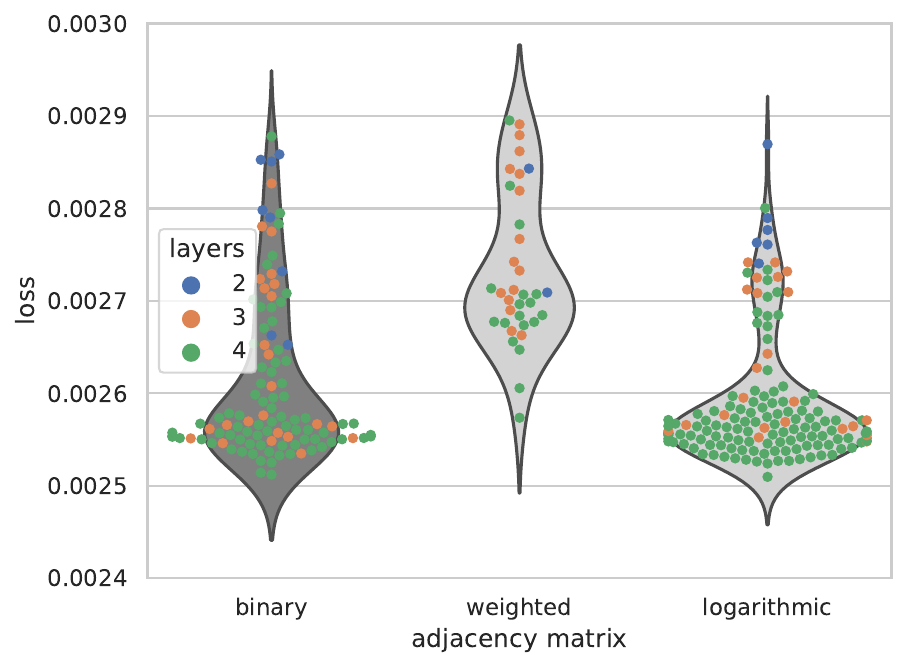}\\
        \footnotesize (a) Observation ratio: $5\%$.
    \end{tabular}
    \begin{tabular}{@{}c@{}}
        \includegraphics[width=\linewidth]{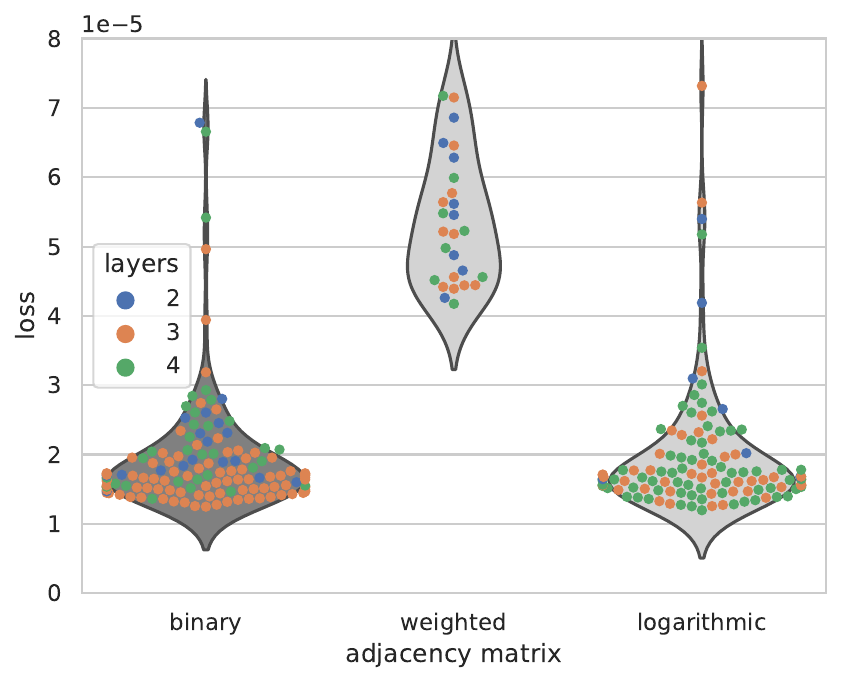}\\
        \footnotesize (b) Observation ratio: $80\%$.
    \end{tabular}
    \caption{Violin plots of the model performances during the hyperparamater optimization at different observation ratios in the Anytown water network.}
    \label{fig:swarm}
\end{figure}

The hyperparameter optimization was carried out for the Anytown WDS with the same observation ratios as the regular trainings.
As the results are similar in all cases, only the results of the lowest and the highest observation ratio are presented in Fig.~\ref{fig:swarm}.
Validation error for all evaluations is depicted in these figures as swarms.
These swarms are separated according to the type of the adjacency matrix used for the calculation of the graph Laplacian.
Finally, violin plots are drawn in the background in order to give a better view on the distribution of the validation error according to a specific swarm.
Each individual inside the swarms is colored by the number of layers used in the GNN.
Other hyperparameters are not visualized in the figures.

The results prove the feasibility of the considerations of Sec.~\ref{sec:effects}.
The models with the weighted adjacency matrix perform worse than the models with the binary adjacency matrix.
Besides, the models with the logarithmically weighted adjacency matrix perform similarly to those with the binary adjacency.

The number of the hidden layers are emphasized in the swarms to show that good performance can be achieved with shallow models.
While at a small observation ratio -- technically, observing only two nodes in the Anytown WDS -- a deeper artificial neural network performs generally slightly better than a shallow one, the shallow one is superior at a high observation ratio.
The performance of the 3-layered and the  4-layered models were similar over the different observation ratios that are not represented here.

For the above reasons, a 3-layered topology was selected as the base of the GNN for the other two WDSs.
The adjacency matrix was set to binary as it performs best close to the logarithmically weighted type.

\subsection{Accuracy of nodal pressure reconstruction}
\begin{figure}
    \centering
    \includegraphics[width=\linewidth]{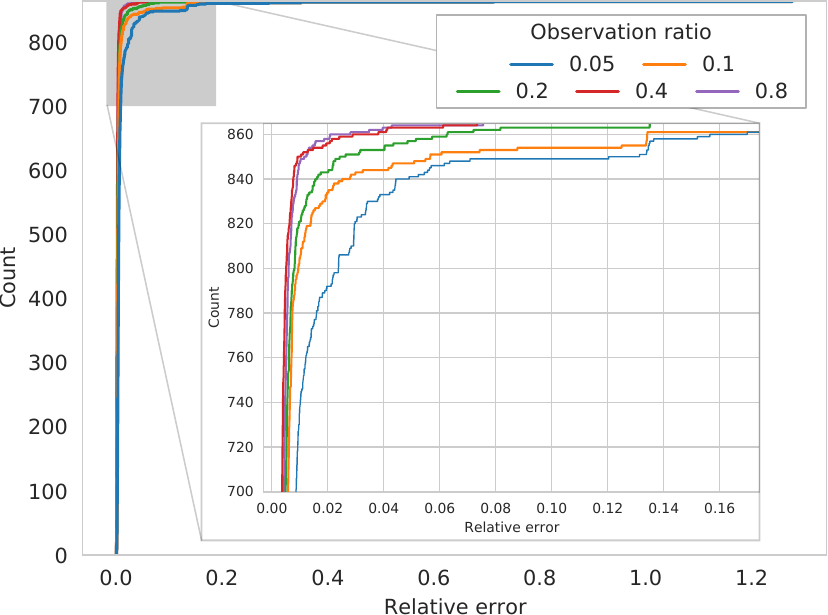}
    \caption{Empirical cumulative distribution function of the relative errors for the nodal pressure in all nodes. Data were calculated on the test scenes of the Richmond water network (containing $865$ nodes) with the $8$th sensor placement (out of $20$) with different observation ratios.}
    \label{fig:ecdf}
\end{figure}

After settling the topologies for every signal-reconstruction model, 20-20 trainings were carried out with sensors installed in different nodes, as described in Sec.~\ref{sec:setup}.
The empirical cumulative distribution function of the relative error with the -- arbitrarily chosen -- $8$th trained model is depicted in Fig.~\ref{fig:ecdf} for the Richmond WDS.
The data were calculated on the test set of scenes.
The relative error is lower than $5\%$ in $95\%$ of the nodes even with the smallest observation ratio and the performance gets better by raising the number of the observed nodes.

\begin{figure*}
    \centering
    \begin{tabular}{@{}c@{}}
        \includegraphics[height=8cm]{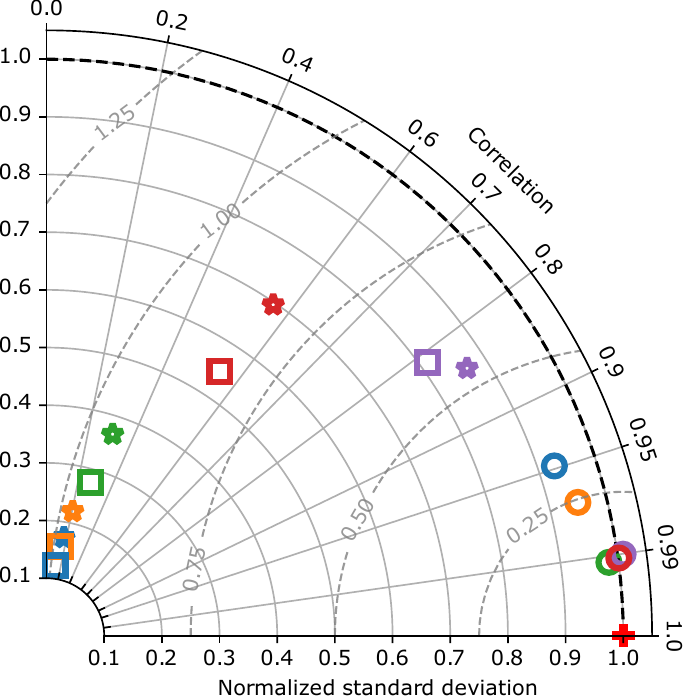} \\
        \footnotesize (a) Anytown
    \end{tabular}
    \begin{tabular}{@{}c@{}}
        \includegraphics[height=8cm]{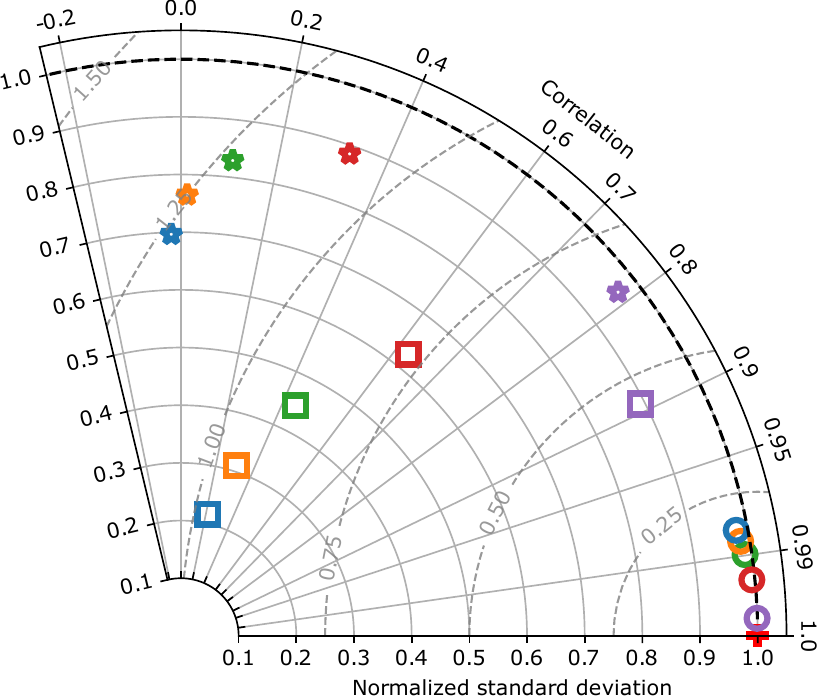} \\
        \footnotesize (b) C-Town
    \end{tabular}
    \begin{tabular}{@{}c@{}}
        \includegraphics[height=5.5cm]{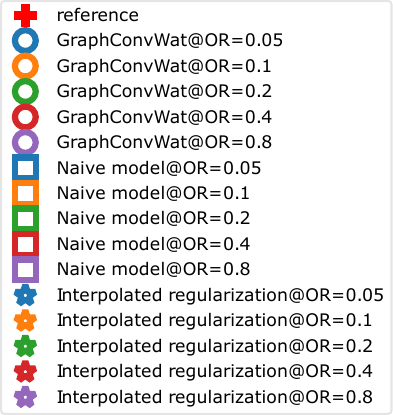} \\
        \footnotesize Legend
    \end{tabular}
    \begin{tabular}{@{}c@{}}
        \includegraphics[height=8cm]{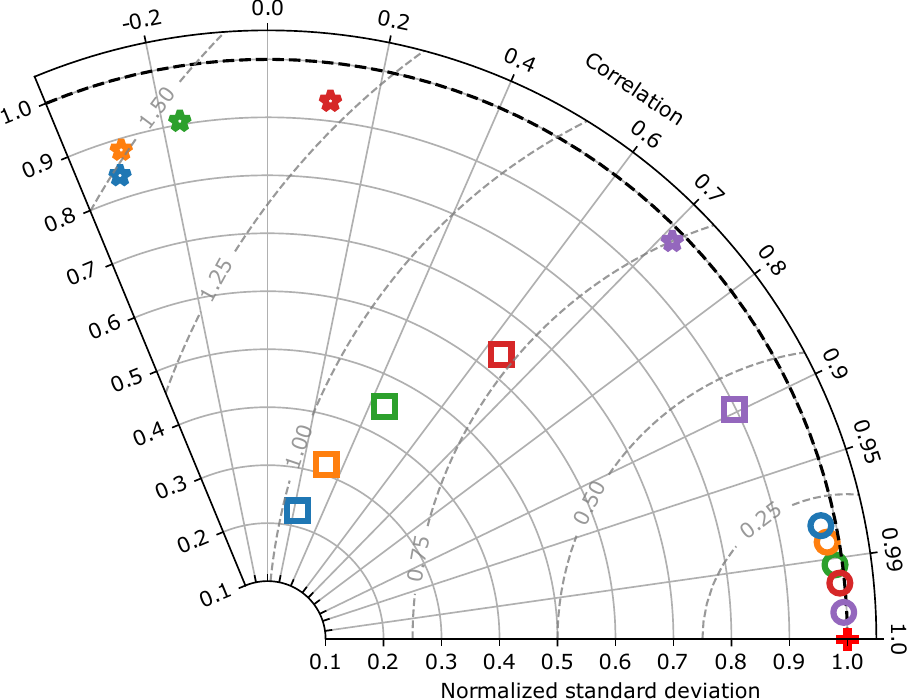} \\
        \footnotesize (c) Richmond
    \end{tabular}
    \caption{Taylor diagram of the reconstruction performance achieved by the na\"ive model, interpolated regularization, and by the graph convolution on water network (ChebNet) at different observation ratios (OR). The test scenes were handled as snapshots of a time-varying field variable and all the nodal pressures -- including the observed ones -- are part of the metrics. The perfect reconstruction is denoted by a red cross as a reference. The markers belonging to ChebNet represents the average metric of the model over $20$ different sensor placements. \textit{Radial coordinate}: standard deviation of the reconstructed nodal pressures normalized with the standard deviation of the ground truth nodal pressures. \textit{Angular coordinate}: correlation coefficient between the reconstructed and the ground truth nodal pressures. \textit{Contours}: centered root mean square error of the reconstruction, the lower the better.}
    \label{fig:taylor-plots}
\end{figure*}

The rest of the cumulative distribution plots are not presented here due to the high number ($300$) of trained models.
An aggregated evaluation is depicted instead in Taylor diagrams \citep{Taylor2001} for the Anytown, C-Town and Richmond WDSs in Fig.~\ref{fig:taylor-plots}(a), (b) and (c), respectively.
The test scenes were handled as snapshots of a time-varying field variable (pressure) and their standard deviation and correlation coefficient compared to the ground truth were calculated.
Three type of metrics are shown in Fig.~\ref{fig:taylor-plots}:
\begin{enumerate}
    \item the normalized standard deviation as the radial coordinate,
    \item the correlation coefficient as the angular coordinate,
    \item the centered root mean square error as contours.
\end{enumerate}
The standard deviation is normalized by the standard deviation of the ground truth pressure field.
The performance of the models with $20$ different sensor placements are averaged per observation ratio to avoid cluttering.

If a model could perfectly reconstruct the nodal pressures in the WDS, it would achieve unit normalized standard deviation and unit correlation coefficient.
As a consequence, its centered root mean square error would be zero.
This place is denoted with a red cross in each figure.
The metrics are computed on the full set of points in the graph, even on the observed ones, thus the points are lying on a half-circular arc with smaller observation ratios tending towards the center of the coordinate system.

Though interpolated regularization is good in maintaining the standard deviation in the recovered signal, it fails to match the ground truth values.
This unfortunate behavior suggests, that the nodal pressure distribution is not smooth enough to meet the assumptions of interpolated regularization.

In contrast, the data-driven approach does not involve any -- possibly inaccurate -- preliminary assumptions on the properties of the pressure distribution, it relies purely on the data.
This inviting property helps ChebNet to clearly outperform both the na\"ive model and the interpolated regularization.
While all the models' performance gets better by raising the observation ratio, the proposed technique achieves a decent performance even with the least observations.
The worst performance is seen in the Anytown water network (Fig.~\ref{fig:taylor-plots}(a)) which has the least nodes among the WDSs.
An observation ratio of $5\%$ means the observation of $2$ nodes only, which explains the sub-average performance.
ChebNet performs better on the other two WDSs, where the centered root mean square error is below $0.25$ for each observation ratio.
The main reason of the deviation from the reference (denoted by a red cross in Fig.~\ref{fig:taylor-plots}) is revealed by the empirical cumulative distribution plot in Fig.~\ref{fig:ecdf}.
While the relative error for the pressure reconstruction is low for most of the nodes (below $5\%$), there are a few nodes with a reconstruction error as high as $100\%$.
Leaving these unrealistic values out of the prediction, the proposed technique is able to reconstruct the vast majority of the unknown signal with an acceptable accuracy from the perspective of WDS operators.

\section{Conclusions}
\label{sec:concs}
A way to utilize graph convolution on water networks has been proposed in the present paper to reconstruct all the nodal pressures from partial observations in water distribution systems.
Several considerations were set up in Sec.~\ref{sec:effects} on how to apply GNNs to this specific task.
Numerical experiments were conducted to examine these considerations and due to the results, the key findings are the following.

Graph neural networks based on the K-localized spectral filtering introduced by \citet{Defferrard2016} are able to reconstruct the nodal pressures on three benchmark WDSs Anytown, C-Town and Richmond with a relative error at most $5\%$ on average with an observation ratio of at least $5\%$.
The reconstruction performance gets better by raising the observation ratio and is superior to interpolated regularization, a common signal recovery method in the literature.

Besides, the performance is adequate even with shallow artificial neural network topologies if the number of filters per layer is kept high enough to model the underlying physics.
The width of the receptive field can be set with the number of hidden layers and the degree of the Chebyshev-polynomials together, the results are satisfying by keeping their combination high enough to span at least the diameter of the graph.

In spite of the inviting possibility to weight the adjacency matrix with hydraulic properties of the WDS, the proposed model gave a similar (or better) reconstruction performance with the binary adjacency matrix utilized.
This induces the use of the binary adjacency matrix over the weighted or the logarithmically weighted one for two reasons:
\begin{enumerate}
    \item the pipe roughness data are often uncertain and as the weighting from the perspective of the information-propagation between nodes is learned during the training, knowledge over the hydraulic weights gives no benefit to the process;
    \item in a well-instrumented WDS where the training data comes from measurement instead of hydraulic simulations, the corresponding information (pipe roughness) is encoded in the training data.
\end{enumerate}

In this setting, the proposed model is able to reconstruct partially observed nodal pressure on WDSs, making it ideal for the management of WDSs, where the number of applicable instruments is limited.
Beside the application in management, the proposed model can open the way for such controller techniques that depend on the full instrumentation of the WDS for optimal operation \citep{Hajgato2020}.

\section*{Acknowledgments}
Bálint Gyires-Tóth is grateful for the financial support of Doctoral Research Scholarship of Ministry of Human Resources (ÚNKP-20-5-BME-210) in the scope of New National Excellence Program, and the János Bolyai Research Scholarship of the Hungarian Academy of Sciences.

The authors thank Yannick Copin for the program code to plot the Taylor diagrams.

\appendix
\section{Software packages and program code}
The graph neural networks were assembled in PyTorch Geometric \citep{Fey2019} based on the PyTorch \citep{Paszke2019} deep learning package.
The hyperparameter optimization was carried out with the Optuna optimization framework \citep{Akiba2019}.
The ample number of hydraulic simulations were carried out with the EPANET solver \citep{Rossman2000} coupled with EPYNET \citep{epynet} to Python.

The program code for the study is available online \citep{graphconvwat}.

\bibliographystyle{elsarticle-harv} 
\bibliography{graco-refs}

\begin{thebibliography}{47}
\expandafter\ifx\csname natexlab\endcsname\relax\def\natexlab#1{#1}\fi
\providecommand{\url}[1]{\texttt{#1}}
\providecommand{\href}[2]{#2}
\providecommand{\path}[1]{#1}
\providecommand{\DOIprefix}{doi:}
\providecommand{\ArXivprefix}{arXiv:}
\providecommand{\URLprefix}{URL: }
\providecommand{\Pubmedprefix}{pmid:}
\providecommand{\doi}[1]{\href{http://dx.doi.org/#1}{\path{#1}}}
\providecommand{\Pubmed}[1]{\href{pmid:#1}{\path{#1}}}
\providecommand{\bibinfo}[2]{#2}
\ifx\xfnm\relax \def\xfnm[#1]{\unskip,\space#1}\fi
\bibitem[{Akiba et~al.(2019)Akiba, Sano, Yanase, Ohta and Koyama}]{Akiba2019}
\bibinfo{author}{Akiba, T.}, \bibinfo{author}{Sano, S.},
  \bibinfo{author}{Yanase, T.}, \bibinfo{author}{Ohta, T.},
  \bibinfo{author}{Koyama, M.}, \bibinfo{year}{2019}.
\newblock \bibinfo{title}{Optuna: A next-generation hyperparameter optimization
  framework}, in: \bibinfo{booktitle}{Proceedings of the 25rd {ACM} {SIGKDD}
  International Conference on Knowledge Discovery and Data Mining}, pp.
  \bibinfo{pages}{2623--2631}.
\bibitem[{Bagloee et~al.(2018)Bagloee, Asadi and Patriksson}]{Bagloee2018}
\bibinfo{author}{Bagloee, S.A.}, \bibinfo{author}{Asadi, M.},
  \bibinfo{author}{Patriksson, M.}, \bibinfo{year}{2018}.
\newblock \bibinfo{title}{Minimization of water pumps' electricity usage: A
  hybrid approach of regression models with optimization}.
\newblock \bibinfo{journal}{Expert Systems with Applications}
  \bibinfo{volume}{107}, \bibinfo{pages}{222--242}.
\newblock \DOIprefix\doi{https://doi.org/10.1016/j.eswa.2018.04.027}.
\bibitem[{van~der Beek(2021)}]{Garzon2021}
\bibinfo{editor}{van~der Beek, P.} (Ed.), \bibinfo{year}{2021}.
\newblock \bibinfo{title}{Modeling Water Distribution Systems with Graph Neural
  Networks}, \bibinfo{organization}{Copernicus Meetings}.
  \bibinfo{publisher}{Copernicus {GmbH}}.
\newblock \DOIprefix\doi{10.5194/egusphere-egu21-9378}.
\bibitem[{Belkin et~al.(2004)Belkin, Matveeva and Niyogi}]{Belkin2004}
\bibinfo{author}{Belkin, M.}, \bibinfo{author}{Matveeva, I.},
  \bibinfo{author}{Niyogi, P.}, \bibinfo{year}{2004}.
\newblock \bibinfo{title}{Regularization and semi-supervised learning on large
  graphs}, in: \bibinfo{booktitle}{Learning Theory}.
  \bibinfo{publisher}{Springer Berlin Heidelberg}, pp.
  \bibinfo{pages}{624--638}.
\newblock \DOIprefix\doi{10.1007/978-3-540-27819-1_43}.
\bibitem[{Bergstra et~al.(2011)Bergstra, Bardenet, Bengio and
  K\'{e}gl}]{Bergstra2011}
\bibinfo{author}{Bergstra, J.}, \bibinfo{author}{Bardenet, R.},
  \bibinfo{author}{Bengio, Y.}, \bibinfo{author}{K\'{e}gl, B.},
  \bibinfo{year}{2011}.
\newblock \bibinfo{title}{Algorithms for hyper-parameter optimization}, in:
  \bibinfo{editor}{Shawe-Taylor, J.}, \bibinfo{editor}{Zemel, R.},
  \bibinfo{editor}{Bartlett, P.}, \bibinfo{editor}{Pereira, F.},
  \bibinfo{editor}{Weinberger, K.Q.} (Eds.), \bibinfo{booktitle}{Advances in
  Neural Information Processing Systems}, \bibinfo{publisher}{Curran
  Associates, Inc.}. pp. \bibinfo{pages}{2546--2554}.
\bibitem[{Bhattacharya et~al.(2003)Bhattacharya, Lobbrecht and
  Solomatine}]{Bhattacharya2003}
\bibinfo{author}{Bhattacharya, B.}, \bibinfo{author}{Lobbrecht, A.H.},
  \bibinfo{author}{Solomatine, D.P.}, \bibinfo{year}{2003}.
\newblock \bibinfo{title}{Neural networks and reinforcement learning in control
  of water systems}.
\newblock \bibinfo{journal}{Journal of Water Resources Planning and Management}
  \bibinfo{volume}{129}, \bibinfo{pages}{458--465}.
\newblock \DOIprefix\doi{10.1061/(asce)0733-9496(2003)129:6(458)}.
\bibitem[{Candelieri et~al.(2020)Candelieri, Galuzzi, Giordani and
  Archetti}]{Candelieri2020}
\bibinfo{author}{Candelieri, A.}, \bibinfo{author}{Galuzzi, B.},
  \bibinfo{author}{Giordani, I.}, \bibinfo{author}{Archetti, F.},
  \bibinfo{year}{2020}.
\newblock \bibinfo{title}{Learning optimal control of water distribution
  networks through sequential model-based optimization}, in:
  \bibinfo{editor}{Kotsireas, I.S.}, \bibinfo{editor}{Pardalos, P.M.} (Eds.),
  \bibinfo{booktitle}{Learning and Intelligent Optimization},
  \bibinfo{publisher}{Springer International Publishing},
  \bibinfo{address}{Cham}. pp. \bibinfo{pages}{303--315}.
\bibitem[{Christensen et~al.(2000)Christensen, Locher and
  Swamee}]{Christensen2000}
\bibinfo{author}{Christensen, B.A.}, \bibinfo{author}{Locher, F.A.},
  \bibinfo{author}{Swamee, P.K.}, \bibinfo{year}{2000}.
\newblock \bibinfo{title}{Limitations and proper use of the hazen-williams
  equation}.
\newblock \bibinfo{journal}{Journal of Hydraulic Engineering}
  \bibinfo{volume}{126}, \bibinfo{pages}{167--170}.
\newblock \DOIprefix\doi{10.1061/(asce)0733-9429(2000)126:2(167)}.
\bibitem[{Defferrard et~al.(2016)Defferrard, Bresson and
  Vandergheynst}]{Defferrard2016}
\bibinfo{author}{Defferrard, M.}, \bibinfo{author}{Bresson, X.},
  \bibinfo{author}{Vandergheynst, P.}, \bibinfo{year}{2016}.
\newblock \bibinfo{title}{Convolutional neural networks on graphs with fast
  localized spectral filtering}.
\newblock \bibinfo{journal}{Advances in Neural Information Processing Systems
  29 (2016)} \href{http://arxiv.org/abs/http://arxiv.org/abs/1606.09375v3}{{\tt
  arXiv:http://arxiv.org/abs/1606.09375v3}}.
\bibitem[{Dwivedi et~al.(2020)Dwivedi, Joshi, Laurent, Bengio and
  Bresson}]{Dwivedi2020}
\bibinfo{author}{Dwivedi, V.P.}, \bibinfo{author}{Joshi, C.K.},
  \bibinfo{author}{Laurent, T.}, \bibinfo{author}{Bengio, Y.},
  \bibinfo{author}{Bresson, X.}, \bibinfo{year}{2020}.
\newblock \bibinfo{title}{Benchmarking graph neural networks}.
\newblock \href{http://arxiv.org/abs/2003.00982v1}{{\tt arXiv:2003.00982v1}}.
\bibitem[{Farias et~al.(2018)Farias, Puig, Rangel and Flores}]{Farias2018}
\bibinfo{author}{Farias, R.L.}, \bibinfo{author}{Puig, V.},
  \bibinfo{author}{Rangel, H.R.}, \bibinfo{author}{Flores, J.},
  \bibinfo{year}{2018}.
\newblock \bibinfo{title}{Multi-model prediction for demand forecast in water
  distribution networks}.
\newblock \bibinfo{journal}{Energies} \bibinfo{volume}{11},
  \bibinfo{pages}{660}.
\newblock \DOIprefix\doi{10.3390/en11030660}.
\bibitem[{Ferrer-Cid et~al.(2021)Ferrer-Cid, Barcelo-Ordinas and
  Garcia-Vidal}]{FerrerCid2021}
\bibinfo{author}{Ferrer-Cid, P.}, \bibinfo{author}{Barcelo-Ordinas, J.M.},
  \bibinfo{author}{Garcia-Vidal, J.}, \bibinfo{year}{2021}.
\newblock \bibinfo{title}{Graph learning techniques using structured data for
  {IoT} air pollution monitoring platforms}.
\newblock \bibinfo{journal}{{IEEE} Internet of Things Journal}
  \bibinfo{volume}{8}, \bibinfo{pages}{13652--13663}.
\newblock \DOIprefix\doi{10.1109/jiot.2021.3067717}.
\bibitem[{Fey and Lenssen(2019)}]{Fey2019}
\bibinfo{author}{Fey, M.}, \bibinfo{author}{Lenssen, J.E.},
  \bibinfo{year}{2019}.
\newblock \bibinfo{title}{Fast graph representation learning with pytorch
  geometric}.
\newblock \href{http://arxiv.org/abs/1903.02428}{{\tt arXiv:1903.02428}}.
\bibitem[{Glorot and Bengio(2010)}]{Glorot2010}
\bibinfo{author}{Glorot, X.}, \bibinfo{author}{Bengio, Y.},
  \bibinfo{year}{2010}.
\newblock \bibinfo{title}{Understanding the difficulty of training deep
  feedforward neural networks}, in: \bibinfo{editor}{Teh, Y.W.},
  \bibinfo{editor}{Titterington, M.} (Eds.), \bibinfo{booktitle}{Proceedings of
  the Thirteenth International Conference on Artificial Intelligence and
  Statistics}, \bibinfo{publisher}{PMLR}, \bibinfo{address}{Chia Laguna Resort,
  Sardinia, Italy}. pp. \bibinfo{pages}{249--256}.
\newblock \URLprefix \url{http://proceedings.mlr.press/v9/glorot10a.html}.
\bibitem[{Hajgat{\'{o}} et~al.(2021)Hajgat{\'{o}}, Gyires-T{\'{o}}th and
  Pa{\'{a}}l}]{graphconvwat}
\bibinfo{author}{Hajgat{\'{o}}, G.}, \bibinfo{author}{Gyires-T{\'{o}}th, B.},
  \bibinfo{author}{Pa{\'{a}}l, G.}, \bibinfo{year}{2021}.
\newblock \bibinfo{title}{{GraphConvWat}}.
\newblock \URLprefix \url{https://github.com/BME-SmartLab/GraphConvWat}.
\bibitem[{Hajgat{\'{o}} et~al.(2020)Hajgat{\'{o}}, Pa{\'{a}}l and
  Gyires-T{\'{o}}th}]{Hajgato2020}
\bibinfo{author}{Hajgat{\'{o}}, G.}, \bibinfo{author}{Pa{\'{a}}l, G.},
  \bibinfo{author}{Gyires-T{\'{o}}th, B.}, \bibinfo{year}{2020}.
\newblock \bibinfo{title}{Deep reinforcement learning for real-time
  optimization of pumps in water distribution systems}.
\newblock \bibinfo{journal}{Journal of Water Resources Planning and Management}
  \bibinfo{volume}{146}, \bibinfo{pages}{04020079}.
\newblock \DOIprefix\doi{10.1061/(asce)wr.1943-5452.0001287}.
\bibitem[{Heinsbroek(2016)}]{epynet}
\bibinfo{author}{Heinsbroek, A.}, \bibinfo{year}{2016}.
\newblock \bibinfo{title}{{EPYNET}}.
\newblock \URLprefix \url{https://github.com/Vitens/epynet}.
\bibitem[{Hendrycks and Gimpel(2016)}]{Hendrycks2016}
\bibinfo{author}{Hendrycks, D.}, \bibinfo{author}{Gimpel, K.},
  \bibinfo{year}{2016}.
\newblock \bibinfo{title}{Gaussian error linear units (gelus)}.
\newblock \href{http://arxiv.org/abs/1606.08415}{{\tt arXiv:1606.08415}}.
\bibitem[{Herrera et~al.(2010)Herrera, Torgo, Izquierdo and
  P{\'{e}}rez-Garc{\'{\i}}a}]{Herrera2010}
\bibinfo{author}{Herrera, M.}, \bibinfo{author}{Torgo, L.},
  \bibinfo{author}{Izquierdo, J.}, \bibinfo{author}{P{\'{e}}rez-Garc{\'{\i}}a,
  R.}, \bibinfo{year}{2010}.
\newblock \bibinfo{title}{Predictive models for forecasting hourly urban water
  demand}.
\newblock \bibinfo{journal}{Journal of Hydrology} \bibinfo{volume}{387},
  \bibinfo{pages}{141--150}.
\newblock \DOIprefix\doi{10.1016/j.jhydrol.2010.04.005}.
\bibitem[{Hochreiter et~al.(2001)Hochreiter, Bengio, Frasconi and
  Schmidhuber}]{Hochreiter2001}
\bibinfo{author}{Hochreiter, S.}, \bibinfo{author}{Bengio, Y.},
  \bibinfo{author}{Frasconi, P.}, \bibinfo{author}{Schmidhuber, J.},
  \bibinfo{year}{2001}.
\newblock \bibinfo{title}{Gradient flow in recurrent nets: the difficulty of
  learning long-term dependencies}, in: \bibinfo{editor}{Kremer, S.C.},
  \bibinfo{editor}{Kolen, J.F.} (Eds.), \bibinfo{booktitle}{A Field Guide to
  Dynamical Recurrent Neural Networks}. \bibinfo{publisher}{IEEE Press}, pp.
  \bibinfo{pages}{237--243}.
\bibitem[{Jiang et~al.(2020)Jiang, Bigot and Maabout}]{Jiang2020}
\bibinfo{author}{Jiang, Y.}, \bibinfo{author}{Bigot, J.},
  \bibinfo{author}{Maabout, S.}, \bibinfo{year}{2020}.
\newblock \bibinfo{title}{Sensor selection on graphs via data-driven node
  sub-sampling in network time series}.
\newblock \href{http://arxiv.org/abs/2004.11815}{{\tt arXiv:2004.11815}}.
\bibitem[{Kingma and Ba(2015)}]{Kingma2015}
\bibinfo{author}{Kingma, D.P.}, \bibinfo{author}{Ba, J.}, \bibinfo{year}{2015}.
\newblock \bibinfo{title}{Adam: {A} method for stochastic optimization}, in:
  \bibinfo{editor}{Bengio, Y.}, \bibinfo{editor}{LeCun, Y.} (Eds.),
  \bibinfo{booktitle}{3rd International Conference on Learning Representations,
  {ICLR} 2015, San Diego, CA, USA, May 7-9, 2015, Conference Track
  Proceedings}, pp. \bibinfo{pages}{1--15}.
\newblock \URLprefix \url{http://arxiv.org/abs/1412.6980}.
\bibitem[{Kipf and Welling(2017)}]{Kipf2016}
\bibinfo{author}{Kipf, T.N.}, \bibinfo{author}{Welling, M.},
  \bibinfo{year}{2017}.
\newblock \bibinfo{title}{Semi-supervised classification with graph
  convolutional networks}, in: \bibinfo{booktitle}{5th International Conference
  on Learning Representations, {ICLR} 2017, Toulon, France, April 24-26, 2017,
  Conference Track Proceedings}, \bibinfo{publisher}{OpenReview.net}.
  p.~\bibinfo{pages}{14}.
\newblock \URLprefix \url{https://openreview.net/forum?id=SJU4ayYgl}.
\bibitem[{Klise et~al.(2013)Klise, Phillips and Janke}]{Klise2013}
\bibinfo{author}{Klise, K.A.}, \bibinfo{author}{Phillips, C.A.},
  \bibinfo{author}{Janke, R.J.}, \bibinfo{year}{2013}.
\newblock \bibinfo{title}{Two-tiered sensor placement for large water
  distribution network models}.
\newblock \bibinfo{journal}{Journal of Infrastructure Systems}
  \bibinfo{volume}{19}, \bibinfo{pages}{465--473}.
\newblock \DOIprefix\doi{10.1061/(asce)is.1943-555x.0000156}.
\bibitem[{Lee et~al.(2014)Lee, Sarp, Jeon and Kim}]{Lee2014}
\bibinfo{author}{Lee, S.W.}, \bibinfo{author}{Sarp, S.}, \bibinfo{author}{Jeon,
  D.J.}, \bibinfo{author}{Kim, J.H.}, \bibinfo{year}{2014}.
\newblock \bibinfo{title}{Smart water grid: the future water management
  platform}.
\newblock \bibinfo{journal}{Desalination and Water Treatment}
  \bibinfo{volume}{55}, \bibinfo{pages}{339--346}.
\newblock \DOIprefix\doi{10.1080/19443994.2014.917887}.
\bibitem[{Lima et~al.(2017)Lima, Brentan, Manzi and Luvizotto}]{Lima2017}
\bibinfo{author}{Lima, G.M.}, \bibinfo{author}{Brentan, B.M.},
  \bibinfo{author}{Manzi, D.}, \bibinfo{author}{Luvizotto, E.},
  \bibinfo{year}{2017}.
\newblock \bibinfo{title}{Metamodel for nodal pressure estimation at near
  real-time in water distribution systems using artificial neural networks}.
\newblock \bibinfo{journal}{Journal of Hydroinformatics} \bibinfo{volume}{20},
  \bibinfo{pages}{486--496}.
\newblock \DOIprefix\doi{10.2166/hydro.2017.036}.
\bibitem[{Liou(1998)}]{Liou1998}
\bibinfo{author}{Liou, C.P.}, \bibinfo{year}{1998}.
\newblock \bibinfo{title}{Limitations and proper use of the hazen-williams
  equation}.
\newblock \bibinfo{journal}{Journal of Hydraulic Engineering}
  \bibinfo{volume}{124}, \bibinfo{pages}{951--954}.
\newblock \DOIprefix\doi{10.1061/(asce)0733-9429(1998)124:9(951)}.
\bibitem[{Mehta et~al.(1999)Mehta, Ghulman and Gerth}]{Mehta1999}
\bibinfo{author}{Mehta, B.}, \bibinfo{author}{Ghulman, H.},
  \bibinfo{author}{Gerth, R.}, \bibinfo{year}{1999}.
\newblock \bibinfo{title}{A new methodology of using design of experiments as a
  precursor to neural networks for material processing: extrusion die design},
  in: \bibinfo{booktitle}{Proceedings of the Second International Conference on
  Intelligent Processing and Manufacturing of Materials.
  {IPMM}{\textquotesingle}99 (Cat. No.99EX296)}, \bibinfo{publisher}{{IEEE}}.
  pp. \bibinfo{pages}{1151--1156}.
\newblock \DOIprefix\doi{10.1109/ipmm.1999.791541}.
\bibitem[{Odan et~al.(2014)Odan, Reis and Kapelan}]{Odan2014}
\bibinfo{author}{Odan, F.K.}, \bibinfo{author}{Reis, L.F.R.},
  \bibinfo{author}{Kapelan, Z.}, \bibinfo{year}{2014}.
\newblock \bibinfo{title}{Use of metamodels in real-time operation of water
  distribution systems}.
\newblock \bibinfo{journal}{Procedia Engineering} \bibinfo{volume}{89},
  \bibinfo{pages}{449--456}.
\newblock \DOIprefix\doi{10.1016/j.proeng.2014.11.211}.
\bibitem[{Paszke et~al.(2019)Paszke, Gross, Massa, Lerer, Bradbury, Chanan,
  Killeen, Lin, Gimelshein, Antiga, Desmaison, Kopf, Yang, DeVito, Raison,
  Tejani, Chilamkurthy, Steiner, Fang, Bai and Chintala}]{Paszke2019}
\bibinfo{author}{Paszke, A.}, \bibinfo{author}{Gross, S.},
  \bibinfo{author}{Massa, F.}, \bibinfo{author}{Lerer, A.},
  \bibinfo{author}{Bradbury, J.}, \bibinfo{author}{Chanan, G.},
  \bibinfo{author}{Killeen, T.}, \bibinfo{author}{Lin, Z.},
  \bibinfo{author}{Gimelshein, N.}, \bibinfo{author}{Antiga, L.},
  \bibinfo{author}{Desmaison, A.}, \bibinfo{author}{Kopf, A.},
  \bibinfo{author}{Yang, E.}, \bibinfo{author}{DeVito, Z.},
  \bibinfo{author}{Raison, M.}, \bibinfo{author}{Tejani, A.},
  \bibinfo{author}{Chilamkurthy, S.}, \bibinfo{author}{Steiner, B.},
  \bibinfo{author}{Fang, L.}, \bibinfo{author}{Bai, J.},
  \bibinfo{author}{Chintala, S.}, \bibinfo{year}{2019}.
\newblock \bibinfo{title}{Pytorch: An imperative style, high-performance deep
  learning library}, in: \bibinfo{editor}{Wallach, H.},
  \bibinfo{editor}{Larochelle, H.}, \bibinfo{editor}{Beygelzimer, A.},
  \bibinfo{editor}{d\textquotesingle Alch\'{e}-Buc, F.}, \bibinfo{editor}{Fox,
  E.}, \bibinfo{editor}{Garnett, R.} (Eds.), \bibinfo{booktitle}{Advances in
  Neural Information Processing Systems}, \bibinfo{publisher}{Curran
  Associates, Inc.}. pp. \bibinfo{pages}{1--12}.
\bibitem[{Romero et~al.(2017)Romero, Ma and Giannakis}]{Romero2017}
\bibinfo{author}{Romero, D.}, \bibinfo{author}{Ma, M.},
  \bibinfo{author}{Giannakis, G.B.}, \bibinfo{year}{2017}.
\newblock \bibinfo{title}{Kernel-based reconstruction of graph signals}.
\newblock \bibinfo{journal}{{IEEE} Transactions on Signal Processing}
  \bibinfo{volume}{65}, \bibinfo{pages}{764--778}.
\newblock \DOIprefix\doi{10.1109/tsp.2016.2620116}.
\bibitem[{Rossman(2000)}]{Rossman2000}
\bibinfo{author}{Rossman, L.A.}, \bibinfo{year}{2000}.
\newblock \bibinfo{title}{EPANET 2 users manual}.
\bibitem[{Shuman et~al.(2013)Shuman, Narang, Frossard, Ortega and
  Vandergheynst}]{Shuman2013}
\bibinfo{author}{Shuman, D.I.}, \bibinfo{author}{Narang, S.K.},
  \bibinfo{author}{Frossard, P.}, \bibinfo{author}{Ortega, A.},
  \bibinfo{author}{Vandergheynst, P.}, \bibinfo{year}{2013}.
\newblock \bibinfo{title}{The emerging field of signal processing on graphs:
  Extending high-dimensional data analysis to networks and other irregular
  domains}.
\newblock \bibinfo{journal}{{IEEE} Signal Processing Magazine}
  \bibinfo{volume}{30}, \bibinfo{pages}{83--98}.
\newblock \URLprefix \url{https://arxiv.org/abs/1211.0053},
  \DOIprefix\doi{10.1109/MSP.2012.2235192},
  \href{http://arxiv.org/abs/http://arxiv.org/abs/1211.0053v2}{{\tt
  arXiv:http://arxiv.org/abs/1211.0053v2}}.
\bibitem[{Sitzenfrei et~al.(2013)Sitzenfrei, Möderl and
  Rauch}]{Sitzenfrei2013}
\bibinfo{author}{Sitzenfrei, R.}, \bibinfo{author}{Möderl, M.},
  \bibinfo{author}{Rauch, W.}, \bibinfo{year}{2013}.
\newblock \bibinfo{title}{Automatic generation of water distribution systems
  based on {GIS} data}.
\newblock \bibinfo{journal}{Environmental Modelling {\&} Software}
  \bibinfo{volume}{47}, \bibinfo{pages}{138--147}.
\newblock \DOIprefix\doi{10.1016/j.envsoft.2013.05.006}.
\bibitem[{Tanaka et~al.(2020)Tanaka, Eldar, Ortega and Cheung}]{Tanaka2020}
\bibinfo{author}{Tanaka, Y.}, \bibinfo{author}{Eldar, Y.C.},
  \bibinfo{author}{Ortega, A.}, \bibinfo{author}{Cheung, G.},
  \bibinfo{year}{2020}.
\newblock \bibinfo{title}{Sampling signals on graphs: From theory to
  applications}.
\newblock \bibinfo{journal}{{IEEE} Signal Processing Magazine}
  \bibinfo{volume}{37}, \bibinfo{pages}{14--30}.
\newblock \DOIprefix\doi{10.1109/msp.2020.3016908}.
\bibitem[{Taormina and Galelli(2018)}]{Taormina2018}
\bibinfo{author}{Taormina, R.}, \bibinfo{author}{Galelli, S.},
  \bibinfo{year}{2018}.
\newblock \bibinfo{title}{Deep-learning approach to the detection and
  localization of cyber-physical attacks on water distribution systems}.
\newblock \bibinfo{journal}{Journal of Water Resources Planning and Management}
  \bibinfo{volume}{144}, \bibinfo{pages}{04018065}.
\newblock \DOIprefix\doi{10.1061/(asce)wr.1943-5452.0000983}.
\bibitem[{Taormina et~al.(2018)Taormina, Galelli, Tippenhauer, Salomons,
  Ostfeld, Eliades, Aghashahi, Sundararajan, Pourahmadi, Banks, Brentan,
  Campbell, Lima, Manzi, Ayala-Cabrera, Herrera, Montalvo, Izquierdo,
  Luvizotto, Chandy, Rasekh, Barker, Campbell, Shafiee, Giacomoni, Gatsis,
  Taha, Abokifa, Haddad, Lo, Biswas, Pasha, Kc, Somasundaram, Housh and
  Ohar}]{Taormina2018a}
\bibinfo{author}{Taormina, R.}, \bibinfo{author}{Galelli, S.},
  \bibinfo{author}{Tippenhauer, N.O.}, \bibinfo{author}{Salomons, E.},
  \bibinfo{author}{Ostfeld, A.}, \bibinfo{author}{Eliades, D.G.},
  \bibinfo{author}{Aghashahi, M.}, \bibinfo{author}{Sundararajan, R.},
  \bibinfo{author}{Pourahmadi, M.}, \bibinfo{author}{Banks, M.K.},
  \bibinfo{author}{Brentan, B.M.}, \bibinfo{author}{Campbell, E.},
  \bibinfo{author}{Lima, G.}, \bibinfo{author}{Manzi, D.},
  \bibinfo{author}{Ayala-Cabrera, D.}, \bibinfo{author}{Herrera, M.},
  \bibinfo{author}{Montalvo, I.}, \bibinfo{author}{Izquierdo, J.},
  \bibinfo{author}{Luvizotto, E.}, \bibinfo{author}{Chandy, S.E.},
  \bibinfo{author}{Rasekh, A.}, \bibinfo{author}{Barker, Z.A.},
  \bibinfo{author}{Campbell, B.}, \bibinfo{author}{Shafiee, M.E.},
  \bibinfo{author}{Giacomoni, M.}, \bibinfo{author}{Gatsis, N.},
  \bibinfo{author}{Taha, A.}, \bibinfo{author}{Abokifa, A.A.},
  \bibinfo{author}{Haddad, K.}, \bibinfo{author}{Lo, C.S.},
  \bibinfo{author}{Biswas, P.}, \bibinfo{author}{Pasha, M.F.K.},
  \bibinfo{author}{Kc, B.}, \bibinfo{author}{Somasundaram, S.L.},
  \bibinfo{author}{Housh, M.}, \bibinfo{author}{Ohar, Z.},
  \bibinfo{year}{2018}.
\newblock \bibinfo{title}{Battle of the attack detection algorithms: Disclosing
  cyber attacks on water distribution networks}.
\newblock \bibinfo{journal}{Journal of Water Resources Planning and Management}
  \bibinfo{volume}{144}, \bibinfo{pages}{04018048}.
\newblock \DOIprefix\doi{10.1061/(asce)wr.1943-5452.0000969}.
\bibitem[{Taylor(2001)}]{Taylor2001}
\bibinfo{author}{Taylor, K.E.}, \bibinfo{year}{2001}.
\newblock \bibinfo{title}{Summarizing multiple aspects of model performance in
  a single diagram}.
\newblock \bibinfo{journal}{Journal of Geophysical Research: Atmospheres}
  \bibinfo{volume}{106}, \bibinfo{pages}{7183--7192}.
\newblock \DOIprefix\doi{10.1029/2000jd900719}.
\bibitem[{Tsiami and Makropoulos(2021)}]{Tsiami2021}
\bibinfo{author}{Tsiami, L.}, \bibinfo{author}{Makropoulos, C.},
  \bibinfo{year}{2021}.
\newblock \bibinfo{title}{Cyber{\textemdash}physical attack detection in water
  distribution systems with temporal graph convolutional neural networks}.
\newblock \bibinfo{journal}{Water} \bibinfo{volume}{13}, \bibinfo{pages}{1247}.
\newblock \DOIprefix\doi{10.3390/w13091247}.
\bibitem[{University of Exeter()}]{cwsExeter}
University of Exeter, \bibinfo{year}{1986}.
\newblock \bibinfo{title}{{C}entre for {W}ater {S}ystems}.
\newblock \URLprefix \url{http://emps.exeter.ac.uk/engineering/research/cws/}.
\bibitem[{Venkitaraman et~al.(2019)Venkitaraman, Chatterjee and
  Handel}]{Venkitaraman2019}
\bibinfo{author}{Venkitaraman, A.}, \bibinfo{author}{Chatterjee, S.},
  \bibinfo{author}{Handel, P.}, \bibinfo{year}{2019}.
\newblock \bibinfo{title}{Predicting graph signals using kernel~regression
  where the input signal is agnostic to a graph}.
\newblock \bibinfo{journal}{{IEEE} Transactions on Signal and Information
  Processing over Networks} \bibinfo{volume}{5}, \bibinfo{pages}{698--710}.
\newblock \DOIprefix\doi{10.1109/tsipn.2019.2936358}.
\bibitem[{Wei et~al.(2020)Wei, Pagani, Fu, Guymer, Chen, McCann and
  Guo}]{Wei2020}
\bibinfo{author}{Wei, Z.}, \bibinfo{author}{Pagani, A.}, \bibinfo{author}{Fu,
  G.}, \bibinfo{author}{Guymer, I.}, \bibinfo{author}{Chen, W.},
  \bibinfo{author}{McCann, J.}, \bibinfo{author}{Guo, W.},
  \bibinfo{year}{2020}.
\newblock \bibinfo{title}{Optimal sampling of water distribution network
  dynamics using graph fourier transform}.
\newblock \bibinfo{journal}{{IEEE} Transactions on Network Science and
  Engineering} \bibinfo{volume}{7}, \bibinfo{pages}{1570--1582}.
\newblock \DOIprefix\doi{10.1109/tnse.2019.2941834}.
\bibitem[{Wu et~al.(2019)Wu, Souza, Zhang, Fifty, Yu and Weinberger}]{Wu2019}
\bibinfo{author}{Wu, F.}, \bibinfo{author}{Souza, A.}, \bibinfo{author}{Zhang,
  T.}, \bibinfo{author}{Fifty, C.}, \bibinfo{author}{Yu, T.},
  \bibinfo{author}{Weinberger, K.}, \bibinfo{year}{2019}.
\newblock \bibinfo{title}{Simplifying graph convolutional networks}, in:
  \bibinfo{editor}{Chaudhuri, K.}, \bibinfo{editor}{Salakhutdinov, R.} (Eds.),
  \bibinfo{booktitle}{Proceedings of the 36th International Conference on
  Machine Learning}, \bibinfo{publisher}{PMLR}. pp.
  \bibinfo{pages}{6861--6871}.
\newblock \URLprefix \url{http://proceedings.mlr.press/v97/wu19e.html}.
\bibitem[{Wu et~al.(2021)Wu, Pan, Chen, Long, Zhang and Yu}]{Wu2021}
\bibinfo{author}{Wu, Z.}, \bibinfo{author}{Pan, S.}, \bibinfo{author}{Chen,
  F.}, \bibinfo{author}{Long, G.}, \bibinfo{author}{Zhang, C.},
  \bibinfo{author}{Yu, P.S.}, \bibinfo{year}{2021}.
\newblock \bibinfo{title}{A comprehensive survey on graph neural networks}.
\newblock \bibinfo{journal}{{IEEE} Transactions on Neural Networks and Learning
  Systems} \bibinfo{volume}{32}, \bibinfo{pages}{4--24}.
\newblock \URLprefix \url{https://arxiv.org/abs/1901.00596},
  \DOIprefix\doi{10.1109/tnnls.2020.2978386},
  \href{http://arxiv.org/abs/1901.00596}{{\tt arXiv:1901.00596}}.
\bibitem[{Xie et~al.(2017)Xie, Zhou, Hou and Zhang}]{Xie2017}
\bibinfo{author}{Xie, X.}, \bibinfo{author}{Zhou, Q.}, \bibinfo{author}{Hou,
  D.}, \bibinfo{author}{Zhang, H.}, \bibinfo{year}{2017}.
\newblock \bibinfo{title}{Compressed sensing based optimal sensor placement for
  leak localization in water distribution networks}.
\newblock \bibinfo{journal}{Journal of Hydroinformatics} \bibinfo{volume}{20},
  \bibinfo{pages}{1286--1295}.
\newblock \DOIprefix\doi{10.2166/hydro.2017.145}.
\bibitem[{Yazdani and Jeffrey(2011)}]{Yazdani2011}
\bibinfo{author}{Yazdani, A.}, \bibinfo{author}{Jeffrey, P.},
  \bibinfo{year}{2011}.
\newblock \bibinfo{title}{Complex network analysis of water distribution
  systems}.
\newblock \bibinfo{journal}{Chaos: An Interdisciplinary Journal of Nonlinear
  Science} \bibinfo{volume}{21}, \bibinfo{pages}{016111}.
\newblock \DOIprefix\doi{10.1063/1.3540339}.
\bibitem[{Zhou et~al.(2021)Zhou, Jiang, Qian, Ding, Yang and He}]{Zhou2021}
\bibinfo{author}{Zhou, Y.}, \bibinfo{author}{Jiang, J.}, \bibinfo{author}{Qian,
  K.}, \bibinfo{author}{Ding, Y.}, \bibinfo{author}{Yang, S.H.},
  \bibinfo{author}{He, L.}, \bibinfo{year}{2021}.
\newblock \bibinfo{title}{Graph convolutional networks based contamination
  source identification across water distribution networks}.
\newblock \bibinfo{journal}{Process Safety and Environmental Protection}
  \DOIprefix\doi{10.1016/j.psep.2021.09.008}.

\end{thebibliography}

\end{document}